\documentclass[lettersize,journal]{IEEEtran}

\usepackage{amsmath,amsfonts}
\usepackage{algorithmic}
\usepackage{algorithm}
\usepackage{array}
\usepackage{booktabs}
\usepackage{colortbl}
\usepackage{cite}
\usepackage[caption=false,font=normalsize,labelfont=sf,textfont=sf]{subfig}
\usepackage{textcomp}
\usepackage{stfloats}
\usepackage{url}
\usepackage{verbatim}
\usepackage{graphicx}
\usepackage{hyperref} 
\usepackage[table]{xcolor}
\usepackage{multirow}
\usepackage{cleveref}
\Crefname{figure}{Fig.}{Figs.}

\def\ourmodel{FSANet}
\def\ourdataset{SceneX}

\begin{document}

\title{\ourmodel: Frequency-Spatial Aware Network for Image Segmentation}

\author{Ruibo~Wang,
        Ziyi~Shen,
        Huaming~Wu,~\IEEEmembership{Senior Member,~IEEE,}
        Dong~Liang,~\IEEEmembership{Senior Member,~IEEE,}
        and~Kun~Shang%
        \thanks{Ruibo~Wang is with the Faculty of Electrical Engineering, Mathematics and Computer Science, Delft University of Technology, 2600 AA Delft, Netherlands.}%
        \thanks{Ziyi~Shen and Dong~Liang are with the School of Biomedical Engineering, Southern Medical University, Guangzhou, Guangdong, China.}%
        \thanks{Huaming Wu is with the Center for Applied Mathematics, Tianjin University, Tianjin 300072, China.}%
        \thanks{Dong~Liang and Kun~Shang are with the Research Center for Medical AI, Shenzhen Institutes of Advanced Technology, Chinese Academy of Sciences, Shenzhen, Guangdong, China.}%
        \thanks{Corresponding authors: Dong~Liang (dong.liang@siat.ac.cn) and Kun~Shang (kunzzz.shang@gmail.com).}%
        \thanks{Code is available at https://github.com/wrbcode/FSANet.}%
}


\maketitle

\begin{abstract}
Image segmentation remains challenging due to occlusions, poor lighting, and irregular structures. Although transformer-based methods achieve high accuracy, they rely heavily on long-range spatial features, leading to high computational costs and neglecting prior knowledge or noise patterns, resulting in missing details and unclear boundaries. To address these issues, we propose \textbf{F}requency \textbf{S}patial \textbf{A}ware \textbf{Net}work (\ourmodel), which integrates prior knowledge with a dual-domain solver to sequentially adapt to diverse segmentation tasks. Specifically, we design three key modules: (1) Structure Prior Module, which recovers overlooked details; (2) Dual-Domain Awareness Module, which captures salient features while disentangling noise; and (3) Edge Estimation Module, which enhances edge awareness for more precise segmentation. In addition, the limited availability of comprehensive segmentation datasets covering various real-world scenarios hinders the performance of existing methods. To address this, we introduce \ourdataset, a novel open-source dataset featuring 10 challenging non-ideal scenarios, establishing a new benchmark for evaluating and improving the robustness and real-world applicability of the segmentation models. Extensive experiments demonstrate the efficiency and effectiveness of \href{https://github.com/wrbcode/FSANet}{\ourmodel}.
\end{abstract}


\section{Introduction}
\IEEEPARstart{A}{ccurate} object segmentation serves as a fundamental cornerstone for high-level scene understanding, playing a critical role in a wide range of applications, including medical imaging~\cite{lu2024lm}, autonomous driving~\cite{muralidhara2025domain}, and robotic sensing and navigation~\cite{hurtado2022semantic}. Recent advances in deep learning have significantly boosted segmentation performance under controlled conditions. In particular, Convolutional Neural Networks (CNNs)~\cite{ronneberger2015u,elgamily2025novel,cao2024medsegmamba} have driven considerable progress by leveraging hierarchical feature learning. Nonetheless, the intrinsic locality of convolutional operations restricts their capacity to capture long-range dependencies and integrate global semantic contexts. To address this, Transformer-based methods~\cite{xie2021segformer,zheng2021rethinking,chai2025transdeep} have been developed to enhance spatial reasoning and contextual modeling via self-attention mechanisms, albeit at the cost of substantial computational complexity.

\begin{figure}[ht!]
\centering
\includegraphics[width=0.95\linewidth]{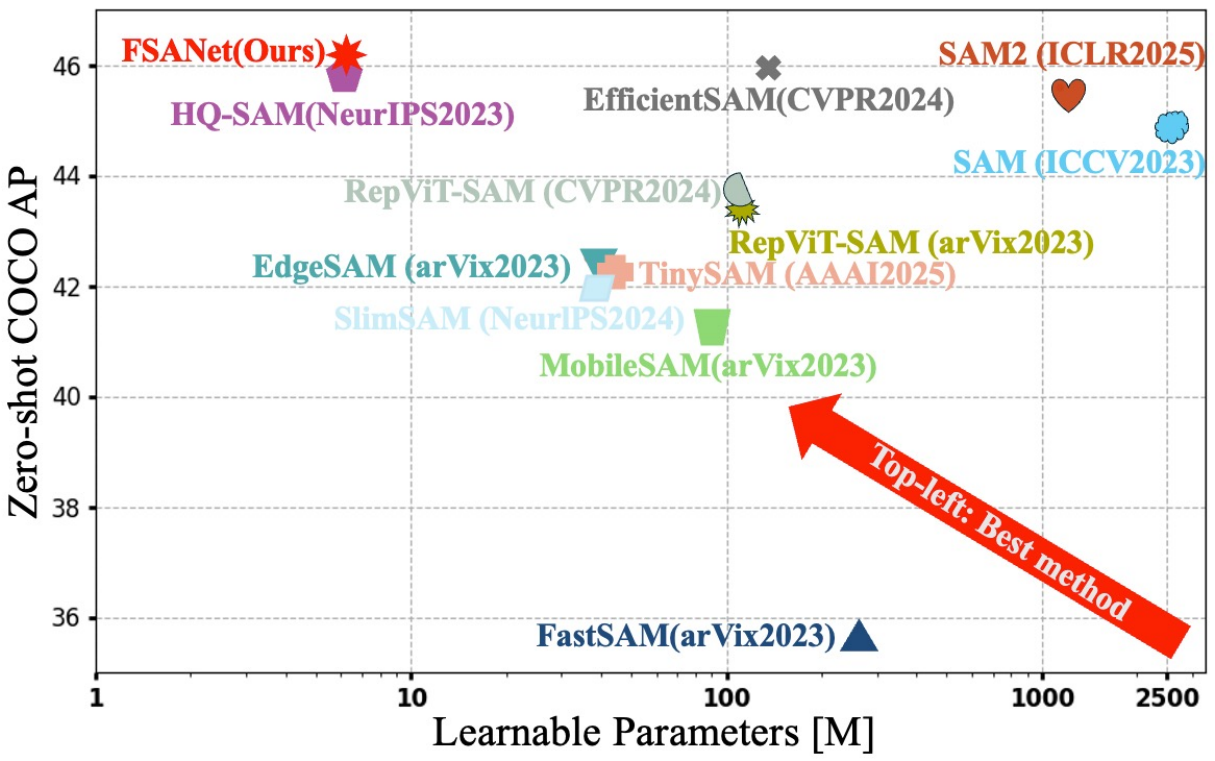}
\caption{Zero-shot Performance vs. Learnable Parameters for an array of SAM variants on the COCO dataset~\cite{lin2014microsoft}.}
\label{fig: compare_variants}
\end{figure}

Real-world settings, by contrast, are inherently dynamic and unpredictable, presenting severe challenges to segmentation models~\cite{Zhou2023Image} designed for controlled environments. These challenges frequently manifest as degraded segmentation quality and error propagation that adversely affects downstream processes such as object tracking~\cite{drayer2016object} and semantic reasoning~\cite{wang2022deep}. Such shortcomings underscore the pressing demand for segmentation frameworks that are both robust and generalizable, capable of adapting to diverse and unconstrained visual scenarios.

To enhance flexibility and generalization across diverse segmentation tasks~\cite{li2024saliency}, unified frameworks such as OneFormer~\cite{jain2023oneformer} and UNINEXT~\cite{yan2023universal} have been proposed. These methods aim to integrate multiple segmentation paradigms, such as semantic, instance, and panoptic segmentation, within a single architectural design. While conceptually elegant, their practical generalization remains constrained. Adaptation to new tasks often requires additional fine-tuning or task-specific prompting, limiting their applicability in open-world and zero-shot settings. More recently, foundation models including CLIPSeg~\cite{luddecke2022image} and the Segment Anything Model (SAM)~\cite{kirillov2023segment} have shown impressive zero-shot segmentation ability by leveraging large-scale vision-language pretraining. Nevertheless, despite their strong generalization, these models frequently underperform in fine-grained segmentation scenarios involving complex object structures and significant global noise~\cite{zhang2023faster,chen2024slimsam,chen2019visual}. In parallel, a body of multi-scale and frequency-aware methods for degraded imagery---video deraining~\cite{zhang2022enhanced}, snow removal with semantic and depth priors~\cite{zhang2021deep}, Taylor-expanded transformers for image restoration~\cite{jin2025mb}, lightweight super-resolution~\cite{zhao2026echosr}, and condition-guided multi-modal prediction with intention and interaction priors~\cite{liu2025condition}---demonstrates the value of explicit multi-scale and spectral priors under adverse conditions, motivating our frequency--spatial design.

This underperformance can be attributed to three primary factors:
(1)~\textbf{Underutilization of prior knowledge}: Many existing methods~\cite{xiong2024efficientsam,kim2024otseg} heavily rely on spatial features extracted by Vision Transformers (ViTs)~\cite{alexey2020image} while insufficiently leveraging structural priors, resulting in suboptimal performance in detailed structure recovery and precise boundary delineation. (2)~\textbf{Ineffective feature distribution modeling}: As noted by Cong et al.~\cite{cong2024semi}, real-world noise often exhibits highly coupled and frequency-dependent characteristics that are not adequately captured by current segmentation frameworks, limiting their capacity to represent such complex noise patterns. (3)~\textbf{Limited training data diversity}: Although widely adopted benchmarks such as COCO~\cite{lin2014microsoft}, Cityscapes~\cite{cordts2016cityscapes}, and LVIS~\cite{gupta2019lvis} have driven progress in structured and well-illuminated environments, they often lack representation of the full spectrum of real-world conditions, including frequent occlusions, poor lighting, cluttered backgrounds, and domain-specific artifacts. Consequently, the generalization capability of current segmentation models remains constrained in unconstrained settings~\cite{qi2023small,wang2024polyp,chang2024DRNet,wu2024toward}.

In this paper, we propose a Frequency--Spatial Aware Network (\ourmodel) that addresses key limitations of existing segmentation methods~\cite{nguyen2019multi} in handling structural ambiguity, global noise, and boundary degradation. Achieving consistent performance gains over state-of-the-art methods that are most pronounced on the boundary-sensitive mBIoU metric, where they widen monotonically with backbone capacity ($+1.3/{+}1.5/{+}2.4$ average mBIoU from ViT-B to ViT-H, \Cref{tab: compare_sam}), as shown in~\Cref{fig: compare_variants}, \ourmodel\ is a noise-resilient and structure-aware segmentation framework that holistically integrates spatial and frequency-domain information for robust scene understanding. It consists of three complementary components: (1) a \textbf{Structure Prior Module (SPM)} that incorporates structural priors to improve spatial coherence and recover fine-grained object details; (2) a \textbf{Dual Domain Awareness Module (DDAM)} that mitigates frequency-domain noise and enhances discriminative features via Frequency Dynamic Filtering and Lightweight Spatial Enhancement Blocks; and (3) an \textbf{Edge Estimation Module (EEM)} that refines boundary localization using edge-aware loss constraints. To further support real-world segmentation under challenging conditions, we introduce \ourdataset, a novel open-source dataset covering ten non-ideal scenarios designed to advance the development and evaluation of segmentation models in diverse adverse environments. The main contributions of this work are summarized as follows:

\begin{itemize}
\item \textbf{Structural Guidance for Spatial Coherence.} We design a SPM that embeds structural priors to reinforce spatial coherence and restore fine object details. This is complemented by an EEM that preserves boundary integrity through edge-aware constraints, together providing explicit structural guidance and precise object delineation.
\item \textbf{Dual-Domain Modeling for Noise Robustness.} We develop a DDAM that disentangles frequency-coupled noise and amplifies salient cues via frequency-adaptive filtering and lightweight spatial enhancement. This dual-domain modeling significantly improves segmentation robustness under real-world noise.    
\item \textbf{Open Dataset for Challenging Scenarios.} We present \ourdataset, an open-source dataset spanning ten challenging non-ideal scenarios. It supports both model training and systematic evaluation under diverse adverse conditions, fostering progress in real-world segmentation.
\item \textbf{Extensive Experimental Validation.} Through comprehensive experiments on standard benchmarks and real-world data, we demonstrate that \ourmodel\ achieves superior segmentation accuracy and exhibits strong cross-domain generalization capability.
\end{itemize}    

\section{Motivation} \label{motivation}

\begin{figure*}[!t]
\centering
\includegraphics[width=0.9\textwidth]{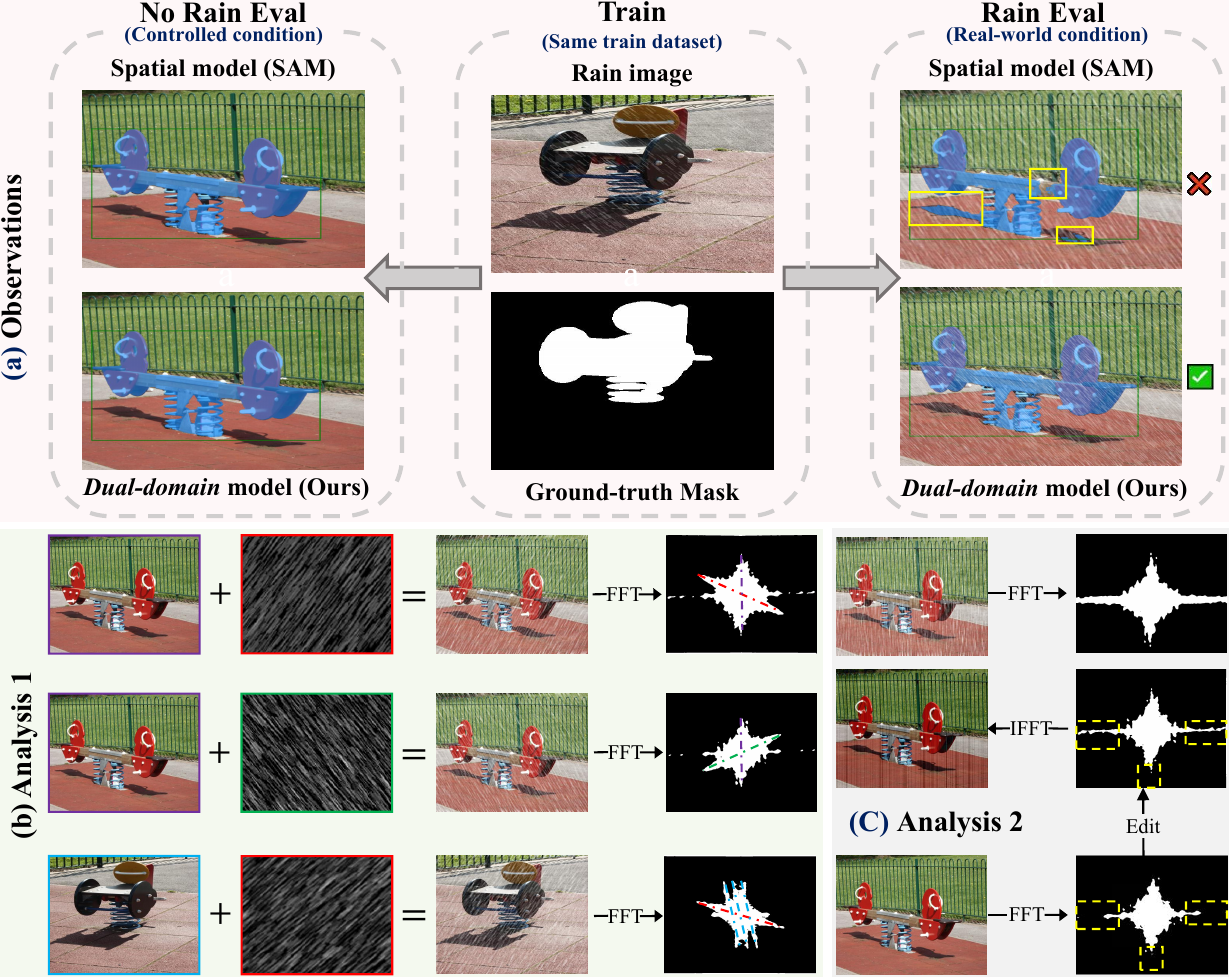}
\caption{\textbf{Motivation for Frequency-Aware Modeling}. (a) \textbf{Robustness Comparison}: Contrast between spatial-only models (SAM) and our dual-domain approach under controlled vs. real-world noisy conditions. (b) \textbf{Spectral Analysis}: Illustration of distinct directional energy distributions for global noise patterns, highlighting their consistency across varying objects. (c) \textbf{Global Governance}: Visualization of global spatial structural distortions induced by local frequency component editing.}
\label{fig: motivation}
\end{figure*}

Understanding the frequency-domain manifestations of real-world noise is essential for building robust segmentation models. As illustrated in \Cref{fig: motivation}, our analysis yields three critical observations:

(a) \textbf{Robustness Dilemma in Spatial Models}: As shown in \Cref{fig: motivation}-(a), while spatial-only models (e.g., SAM) perform accurately in controlled, clean scenarios ("No Rain Eval"), they suffer catastrophic degradation under real-world environmental noise ("Rain Eval"). In contrast, our dual-domain approach maintains high consistency with the ground truth by effectively leveraging frequency priors, demonstrating superior robustness.

(b) \textbf{Spectral Decoupling of Noise and Semantics}: Further spectral analysis in \Cref{fig: motivation}-(b) reveals that global noise patterns possess distinct, disentangleable frequency signatures. Comparing the first two rows, we observe that rain streaks with different spatial orientations manifest as high-magnitude spectral streaks in orthogonal directions in the frequency domain. Crucially, a comparison between the second and third rows demonstrates that this spectral representation remains invariant to the underlying semantic content: when the noise pattern persists, the characteristic "frequency spikes" remain identical even as the scene changes. This confirms that the frequency domain effectively separates noise patterns from semantic information.

(c) \textbf{Global Governance of Frequency Components}: Finally, the frequency manipulation experiment in \Cref{fig: motivation}-(c) highlights the holistic impact of local frequency components. By isolating specific high-frequency regions (highlighted in yellow dashed boxes) associated with noise and performing an Inverse FFT (IFFT), we observe that these local spectral edits induce global, rain-like structural artifacts across the entire reconstructed image. This phenomenon validates that the frequency domain not only characterizes global noise but also governs spatial structural integrity globally.

These findings indicate that frequency-domain analysis can effectively disentangle spatially entangled noise patterns. The translation invariance of the magnitude spectrum supports robust pattern matching, while its global receptive field facilitates the joint modeling of fine-grained anomalies and coarse structures. Consequently, integrating frequency-domain cues serves as a powerful prior for achieving accurate and robust segmentation in unseen, noisy conditions.

To place these observations on a quantitative footing, we measure the radially averaged power spectrum of 100 clean images and of the same images under six controlled degradations. \Cref{tab: spectra} reports the energy of the high band, above half the Nyquist frequency, relative to the clean image. Blur and JPEG compression remove high-frequency energy, whereas additive noise, low light and rain add it, so each degradation carries a distinct and reproducible spectral signature. A single fixed filter cannot serve both regimes, which is why the filter used in \Cref{method} is learned rather than designed.

\begin{table}[!t]
\centering
\caption{Energy of the high-frequency band (above half the Nyquist frequency) under six controlled degradations, relative to the clean image, averaged over 100 images.}
\label{tab: spectra}
\resizebox{\linewidth}{!}{
\begin{tabular}{lcccccc}
\toprule
Degradation & Blur & JPEG & Haze & Rain & Low light & Noise \\
\midrule
High-band energy vs.\ clean & $\times0.04$ & $\times0.64$ & $\times0.86$ & $\times1.24$ & $\times2.17$ & $\times2.71$ \\
\bottomrule
\end{tabular}}
\end{table}

Because \ourmodel\ filters the spectrum of a deep feature map rather than that of the image, we further verify that the two spectra are linked. Passing 250 clean and degraded images through the frozen encoder, the high-band energy of the feature map follows that of the image, but selectively: the loss of high frequencies caused by blur propagates into the feature spectrum (correlation $+0.56$ across the blur sweep), whereas the encoder itself removes most of the high-frequency energy added by noise (correlation $-0.81$). The feature spectrum is thus a structured, degradation-dependent transform of the image spectrum, which is what makes spectral filtering meaningful at the feature level.

\section{Method}\label{method}

\begin{figure*}[ht!]
\centering
\includegraphics[width=0.95\textwidth]{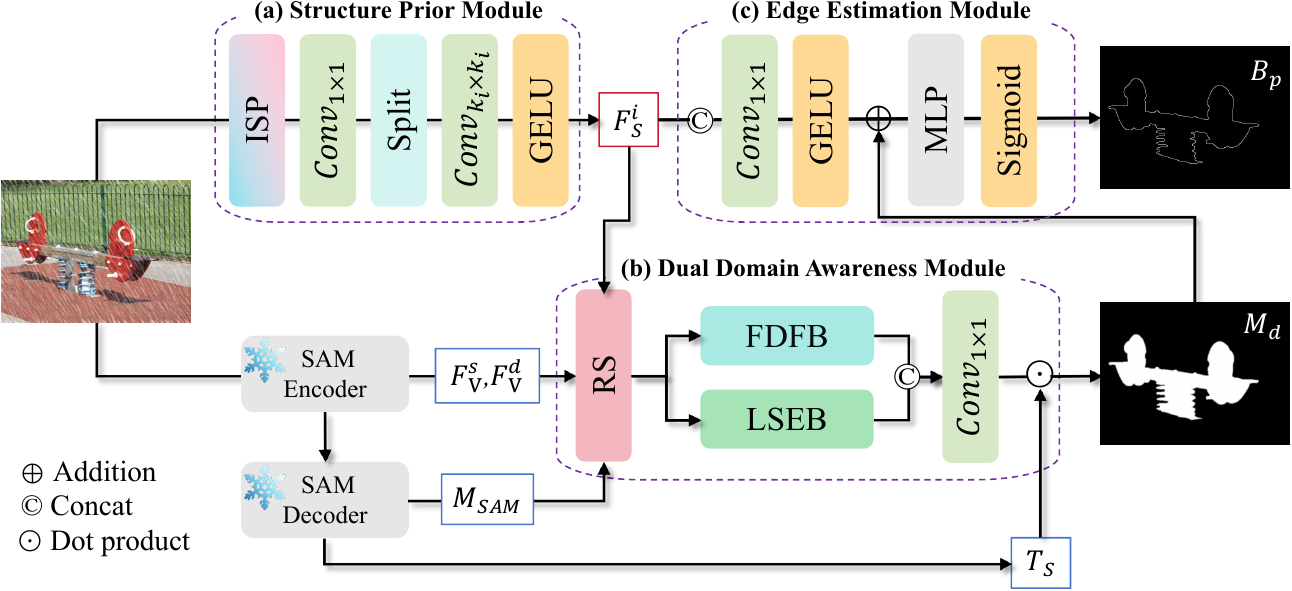}
\caption{Overall architecture of the proposed \ourmodel, which consists of three core components: the Structure Prior Module (SPM) for recovering structural information, the Dual Domain Awareness Module (DDAM) for cross-domain modeling in complex scenarios, and the Edge Estimation Module (EEM) for enhancing boundary awareness. For clarity, the prompt encoder, prompt tokens, and output tokens in SAM are omitted.}
\label{framework}
\end{figure*}

This section elaborates on the architectural components of \ourmodel, as illustrated in \Cref{framework}.

\subsection{Structure Prior Module}
The SPM, depicted in \Cref{framework}(a), integrates object-level structural priors into boundary prediction and enhances the recovery of fine structural details through a gated interaction with ViT features. Specifically, a structural prior is constructed to suppress global noise while preserving essential structural information~\cite{gao2024efficient}. This prior is computed as the channel-wise difference between the maximum and minimum values of the input image $I \in \mathbb{R}^{H\times W \times C}$:
\begin{equation}
I_{S}(h, w) = \mathop{\max}_{c \in \{r,g,b\}} I^c(h, w) - \mathop{\min}_{d \in \{r,g,b\}} I^d(h, w).
\end{equation}
The resulting $I_{S}$ is then processed to extract multi-scale structural representations:
\begin{equation}
F_{S}^{i} = \sigma_g \!\left(C_{k_i\times k_i}\!\left(\mathcal{S}(C_{1\times1}(I_{S})) \right)\right),
\end{equation}
where $ C_{1\times1} $ represents a $1 \times 1 $ convolution, $ C_{k_i\times k_i} $ denotes convolutions with varying kernel sizes $  k_i \in \{7,9\} $, $\mathcal{S}(\cdot)$ is the split operation, $ \sigma_g $ is the GeLU activation function~\cite{hendrycks2016gaussian}, and  $F_{S}^{i}$  corresponds to the feature group indexed by $i \in \{1, 2\} $.

\subsection{Dual Domain Awareness Module}

As illustrated in \Cref{framework}(b), the Dual Domain Awareness Module (DDAM) serves as the core component of \ourmodel. It employs a cross-domain modeling strategy to enhance perceptual robustness under diverse conditions. A Refinement Step (RS) is designed to mitigate the severe degradation of segmentation accuracy in complex real-world environments by suppressing global noise and amplifying semantically meaningful features. This RS process is implemented through an Expansion layer $E(\cdot)$~\cite{sandler2018mobilenetv2} and a Projection layer $P(\cdot)$~\cite{sandler2018mobilenetv2}, and is formulated as:
\begin{equation} 
D = \text{CRA}_m(M_{SAM}+F_{F}) + M_{SAM} + F_{F},
\end{equation}
where $\text{CRA}_m(\cdot)$ refers to a modified Channel Reduction Attention mechanism~\cite{kang2024metaseg}\footnote{In CRA$_m$, the standard pooling operation in CRA is replaced with max-pooling.},  $M_{SAM}$ is the mask feature extracted from the frozen SAM encoder, and $F_{F} = \mathrm{Cat}\left(\sigma_g \!\left(P\!\left(\mathcal{S}\big(\sigma_g(C_{3 \times 3}^{d}(E(F_{V}^s+F_{V}^{d})))\big)\right) \odot F_{S}^{i}\right)\right)$ represents the fused feature derived from the shallow and deep ViT features ($F_{V}^{s}$ and $F_{V}^{d}$) of the frozen SAM encoder. Here, $\mathrm{Cat}$ denotes concatenation, $C_{3 \times 3}^{d}$ denotes a $3 \times 3$ depth-wise convolution, and $\odot$ indicates the dot product.

\subsubsection{Frequency Dynamic Filtering Block}

\begin{figure}[ht!]
    \centering
    \includegraphics[scale=0.6]{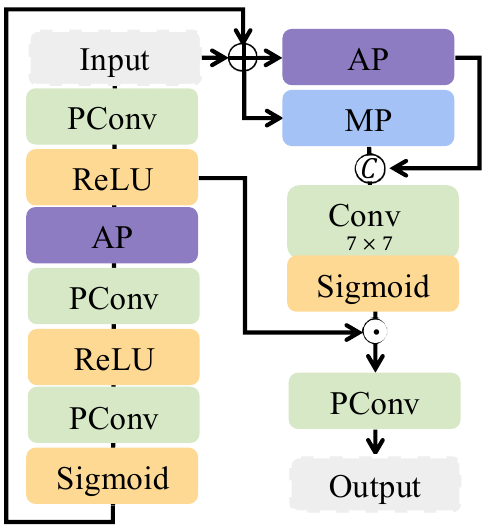}
    \caption{Architecture of the learnable filter $\mathcal{W}$ in the FDFB, where $AP$ and $MP$ denote average pooling and max pooling, respectively. PConv refers to point-wise convolutions.}
    \label{filter}
\end{figure}

The Frequency Dynamic Filtering Block (FDFB) enhances prediction clarity by adaptively disentangling target features from global noise in the frequency domain. Specifically, the Fast Fourier Transform (FFT) $\mathcal{F}(\cdot)$ converts the fused spatial feature representation $D$ into complex frequency features. For each channel $c$, this is expressed as:
\begin{equation} 
Z_c = \mathcal{F}(D_c).
\end{equation}
The corresponding magnitude $A(Z_c)$ and phase $\phi(Z_c)$ are computed as:
\begin{align}
 A(Z_c) &= \sqrt{(\text{Re}(Z_c))^2 + (\text{Im}(Z_c))^2},\\
 \phi(Z_c)&= \operatorname{arctan2}\big(\text{Im}(Z_c),\, \text{Re}(Z_c)\big),
\end{align}
where $\text{Re}(\cdot)$ and $\text{Im}(\cdot)$ denote the real and imaginary parts, respectively.  

The learnable filter $\mathcal{W}$, illustrated in \Cref{filter}, is then applied to the real and imaginary components to enable adaptive separation of target features from irrelevant noise:
\begin{equation} 
[\hat{A}(Z_c), \hat{\phi}(Z_c)] = [\mathcal{W}_A(A(Z_c)) + A(Z_c),  \mathcal{W}_{\phi}(\phi(Z_c)) + \phi(Z_c)],
\end{equation} 
Note that $\mathcal{W}_A$ and $\mathcal{W}_\phi$ share the same architecture but do not share parameters. The cross-enhanced frequency-domain representation is obtained as:
\begin{equation}
  \hat{Z_c}= \hat{A}(Z_c)\cdot \cos(\hat{\phi}(Z_c)) + j\, \hat{A}(Z_c)\cdot \sin(\hat{\phi}(Z_c)),
\end{equation}

where $j$ denotes the imaginary unit, so that $\hat{Z}_c$ is a complex spectrum whose real and imaginary parts are formed from the refined magnitude and phase. The corresponding spatial-domain representation is then reconstructed via inverse FFT:
\begin{equation}
\hat{D}_f= \mathcal{F}^{-1}(\hat{\mathcal{Z}}),
\end{equation}
where $\hat{\mathcal{Z}}=[\hat{Z}_1;\cdots;\hat{Z}_c;\cdots;\hat{Z}_C] \in \mathbb{C}^{H \times (\lfloor W/2\rfloor+1) \times C}$ (the half-spectrum width of the real FFT).

\subsubsection{Lightweight Spatial Enhancement Block}
The Lightweight Spatial Enhancement Block (LSEB) operates in parallel with the frequency-domain processing branch to capture complementary spatial details. By employing SCConv $C_{sc}$~\cite{li2023scconv}, this block efficiently suppresses spatial redundancy. The spatial component $D_s$ is computed as:
\begin{equation}
D_s = C_{sc}\!\big(C_{3 \times 3}(C_{sc}(C_{3 \times 3}(D))) + D\big).
\end{equation}

To combine information from both domains, the reconstructed frequency feature $\hat{D}_f$ and the spatial feature $D_s$ are fused via a linear projection and a residual connection, producing the dual-domain feature $D'$:
\begin{equation}
    D' = C_{1 \times 1} \big(\mathrm{Cat}(\hat{D}_f, D_s)\big) + D.
\end{equation}

Finally, the output $M_d$ of DDAM is obtained by modulating the output tokens $T_s \in \mathbb{R}^{5 \times 256}$ from the frozen SAM with the fused feature $D'$ (here $T_s$ collects the five mask-carrying tokens---the four mask tokens together with the appended high-quality token---out of the six tokens produced by the decoder; the remaining IoU token is reserved for mask-quality prediction and is not modulated):
\begin{equation}
    M_d = T_s \odot D',
\end{equation}
where $\odot$ denotes the dot product with broadcasting. Concretely, as in SAM and HQ-SAM, each of the five mask tokens in $T_s$ is first transformed by a hypernetwork MLP and then contracted with the fused per-pixel feature $D'$ along the channel dimension, producing one mask map per token; $\odot$ abbreviates this token-to-mask operation, which differs from the element-wise product used earlier. Together, these components enable \ourmodel\ to effectively learn category-specific noise features in real-world scenarios, as discussed in~\cref{motivation}.

\subsection{Edge Estimation Module}
The Edge Estimation Module (EEM) is designed to improve boundary segmentation accuracy. It generates the final predicted boundary mask $B_{p}$ by integrating outputs from SPM, and DDAM. The formulation is given by:
\begin{equation}
B_{p} = \sigma_{s} (\mathrm{MLP}(M_d+ \sigma_{g}(C_{1 \times 1}(\mathrm{Cat}(F_{S}^{i})))),
\end{equation}
where $\sigma_{s}$ denotes the sigmoid function, $\mathrm{MLP}$ represents a Multilayer Perceptron.

\subsection{Loss Function}
Accurate segmentation requires not only precise pixel-wise classification but also sharp delineation of object boundaries, particularly in complex or noisy scenes. To this end, we adopt a composite loss function that balances global prediction accuracy with explicit boundary refinement. Specifically, we employ Binary Cross-Entropy (BCE) loss $\mathcal{L}_\text{BCE}$~\cite{hinton2006fast} to supervise pixel-level classification, and Dice loss $\mathcal{L}_\text{Dice}$~\cite{sudre2017generalised} to mitigate class imbalance and enhance region-level consistency. Despite their effectiveness, these losses often provide insufficient constraints on boundary localization, leading to blurred or structurally inconsistent edges~\cite{ke2023segment}.

To explicitly enforce accurate boundary localization, we introduce a Boundary Loss term, defined as:
\begin{equation} 
\begin{split} 
\mathcal{L}_{\text{Boundary}} = & - \frac{1}{M} \sum_{i=1}^M \Big[ M_G^b[i] \cdot \log(M_P^b[i]) \\
                               & + \big(1 - M_G^b[i]\big) \cdot \log\big(1 - M_P^b[i]\big) \Big], 
\end{split} 
\end{equation}
where $M_P^b[i]$ represents the predicted value of the $i$-th pixel in the boundary mask, $M_G^b[i]$ denotes the corresponding ground truth value, and $M$ is the total number of boundary pixels. 

The overall training objective for \ourmodel\ is formulated as a weighted combination of the three loss components:
\begin{equation}
\mathcal{L} = \mathcal{L}_{\text{BCE}} + \alpha \cdot \mathcal{L}_{\text{Dice}} + \beta \cdot \mathcal{L}_{\text{Boundary}}, 
\end{equation}
where $\alpha$ and $\beta$ are hyperparameters that control the contribution of each term. An ablation study on hyperparameter selection is provided in~\Cref{fig: abla-loss-grid-search}.

\subsection{The \ourdataset~Dataset}\label{ourdataset}

\begin{figure}[ht!]
\begin{minipage}[t]{0.50\textwidth}
\centering
\includegraphics[scale=0.35]{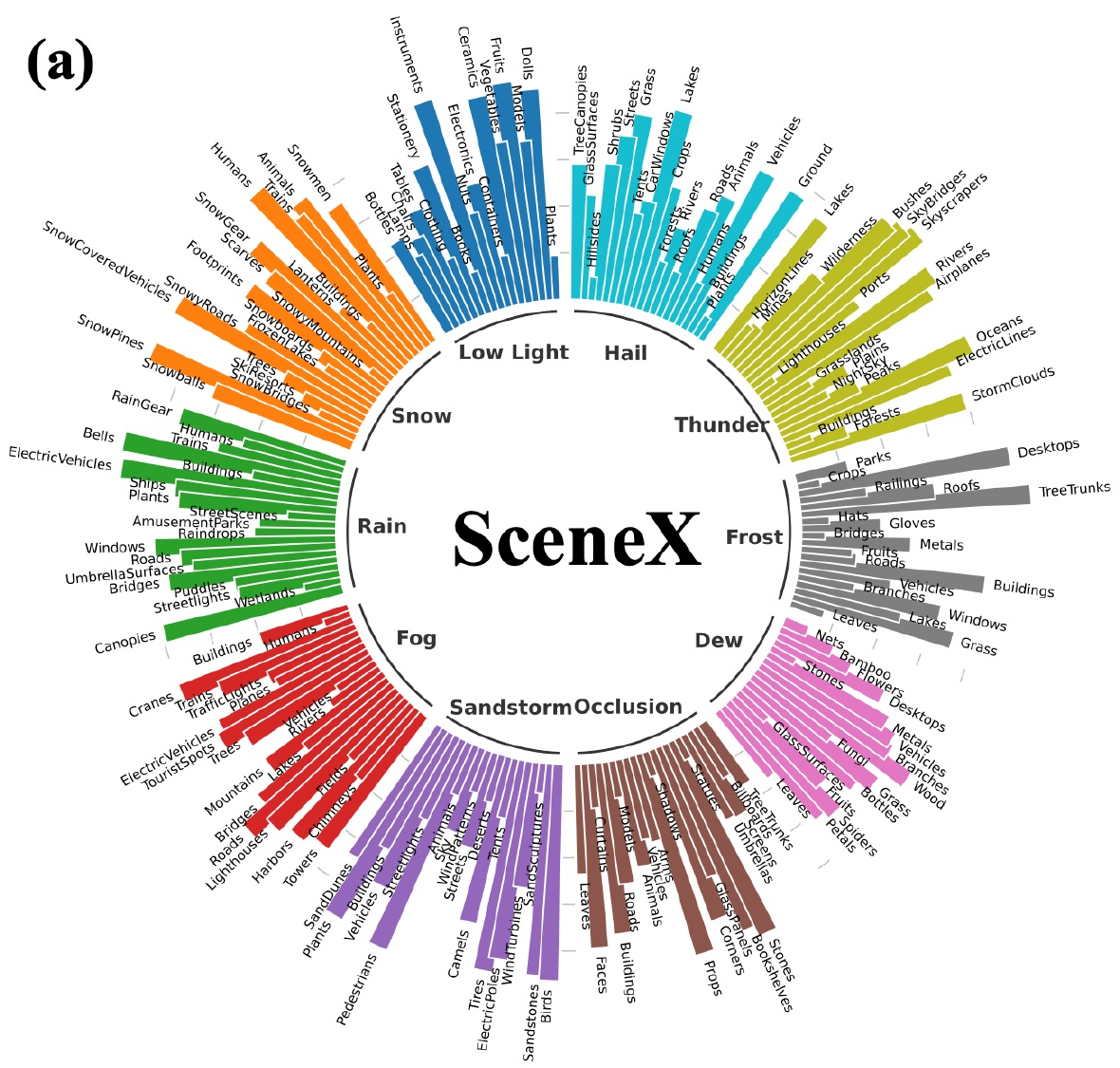}
\end{minipage}
\vspace{0.1cm} 
\begin{minipage}[t]{0.50\textwidth}
\centering
\includegraphics[scale=0.40]{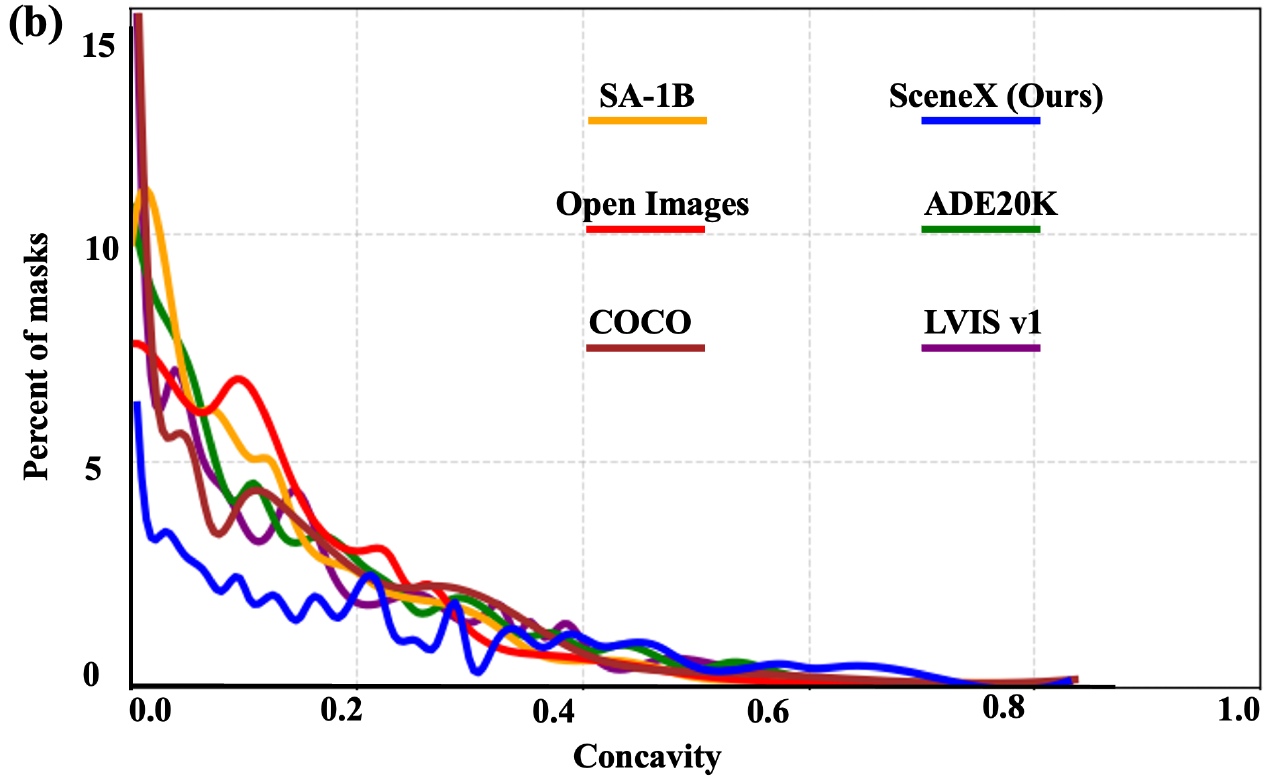}
\end{minipage}
\caption{(a) Major category distribution and representative subcategory examples from \ourdataset. (b) Concavity analysis comparing \ourdataset\ with existing datasets. The X-axis represents Concavity, measuring mask shape complexity (higher values indicate more irregular boundaries). The Y-axis shows the Percent of Masks, indicating the proportion of masks at each concavity level.}
\label{fig: dataset}
\end{figure}

To facilitate the development of robust segmentation models, we introduce \ourdataset, a novel open-source dataset comprising 11,000 high-quality images collected from ten diverse real-world scenarios. As illustrated in \Cref{fig: dataset} (a), this dataset provides a comprehensive resource for training advanced segmentation models. The images are sourced from both online repositories and original captures, ensuring broad scene diversity and real-world relevance~\cite{shi2024zero, liao2018hdpnet}. The dataset encompasses substantial variations in lighting conditions, camera perspectives, and weather effects, making it particularly suitable for developing adaptable and noise-resistant segmentation models.

\textbf{Image Collection}: Image resolutions in \ourdataset\ range from $200 \times 125$ to $4000 \times 3000$, with an average resolution of $1041 \times 732$. Compared to widely used benchmarks such as DUTS~\cite{wang2017learning} ($372 \times 322$) and COIFT~\cite{liew2021deep} ($600 \times 488$), \ourdataset\ offers significantly higher resolution, thereby facilitating more effective multi-scale learning for segmentation models operating at varying levels of detail. 

\textbf{Annotation Methodology}: We employ a semi-automated annotation pipeline to ensure high-quality mask generation. Initial object localization is performed using saliency detection techniques, followed by segmentation using five representative methods: EdgeSAM~\cite{zhou2023edgesam}, EfficientSAM~\cite{xiong2024efficientsam}, SAM~\cite{kirillov2023segment}, HQ-SAM~\cite{ke2023segment}, and TransDeep~\cite{chai2025transdeep}. From the generated candidates, we select the mask with the highest Intersection over Union (IoU) score. Subsequent refinements are applied using edge detection tools from OpenCV~\cite{bradski2000opencv}, followed by manual validation to ensure annotation accuracy. This hybrid approach combining semi-automated techniques with human verification significantly enhances mask quality, producing precise and reliable segmentation annotations.

\textbf{Annotation Reliability}: Because the pipeline is seeded by segmentation models, we verify that the resulting masks do not inherit a bias towards strong image gradients. Using 250 masks per dataset and model-free shape statistics, we compare \ourdataset\ with three datasets whose masks are drawn manually by experts, DIS5K-VD~\cite{qin2022highly}, ThinObject5K-TE~\cite{liew2021deep} and COIFT~\cite{liew2021deep}. \ourdataset\ has the lowest boundary-gradient ratio of the four ($2.95$ against $3.55$, $3.64$ and $6.36$), so its boundaries follow image gradients no more closely than manually drawn ones do, and its thin-structure ratio ($0.182$) lies inside the range of the manual sets ($0.168$ to $0.369$). Its masks are topologically simpler, carrying fewer holes per mask, which reflects both the pipeline and the fact that DIS5K and ThinObject5K are built around intricate objects. We release the automatic and the manually refined subsets so that annotator agreement can be measured directly.

\textbf{Mask Properties}: A distinctive property of \ourdataset\ is the geometric complexity of its masks, quantified by concavity\footnote{Defined as 1 minus mask area divided by area of mask's convex hull~\cite{kirillov2023segment}.}. As shown in \Cref{fig: dataset} (b), a substantial proportion of masks exhibit concavity values between 0.6 and 0.8, indicating prevalent complex shapes and irregular boundaries. This characteristic is crucial for training models to handle intricate object structures. To maintain balanced object size distribution, we categorize masks into size-based bins and perform stratified sampling within each bin. This strategy prevents bias toward specific scales and enables \ourdataset~to effectively support the training of segmentation models across a wide spectrum of object sizes.

\section{Experiments}

\subsection{Experimental Settings}
\textbf{Datasets.} 
Following established evaluation protocols~\cite{kirillov2023segment, ke2023segment}, we conduct comprehensive experiments across ten publicly available datasets characterized by high-quality, fine-grained annotations (averaging 7.4k masks per dataset across over 1,000 diverse categories). To ensure robustness and broad generalization, our training set integrates the training split of our proposed \ourdataset~with six widely adopted benchmarks: DIS5K-TR~\cite{qin2022highly} for objects with varying structural complexities; ThinObject5K-TR~\cite{liew2021deep} for fine, elongated structures; FSS-1000~\cite{li2020fss} for diverse and previously unseen classes; ECSSD~\cite{shi2015hierarchical} for objects in structurally complex backgrounds; MSRA-10K~\cite{cheng2014global} for large-scale pixel-level saliency; and DUTS~\cite{wang2017learning} for challenging scenarios. Here HQSeg-44K denotes the high-quality benchmark suite introduced by HQ-SAM~\cite{ke2023segment}, on whose held-out split \Cref{tab: abla-filter} reports results---disjoint from the data used for training. For evaluation, we utilize the validation split of \ourdataset, DIS5K-VD~\cite{qin2022highly}, and ThinObject5K-TE~\cite{liew2021deep}, complemented by COIFT~\cite{liew2021deep} and HR-SOD~\cite{zeng2019towards} to specifically address the challenges of thin object segmentation in heterogeneous backgrounds and high-resolution saliency detection, respectively. Furthermore, we assess zero-shot generalization on the COCO benchmark~\cite{lin2014microsoft} for common objects and the SGinW~\cite{zou2023generalized} benchmark to rigorously test adaptability in open-world scenarios.

\textbf{Data Processing}~The data preprocessing pipeline plays a critical role in ensuring that the model is exposed to a diverse range of inputs and is capable of generalizing across various scenarios. We begin by loading the data using separate data loaders for training and validation, allowing for efficient batching and transformation. This separation ensures that the training data is properly utilized while preventing any data leakage into the validation phase. 
For data augmentation, we apply a series of transformations to the training set to enhance the model's robustness to variations in object scale, orientation, and background complexity. Specifically, we apply random flipping with a probability of 50\% and large-scale jittering, which randomly varies the scale between 0.1 and 2. These transformations help the model adapt to different object scales and orientations, improving its ability to generalize to unseen data.
In addition to geometric augmentations, we compute an Image Structure Prior (ISP) for each image based on its maximum and minimum pixel values. This step, inspired by recent work~\cite{gao2024efficient}, enhances the model's robustness against complex backgrounds by providing a normalized representation of the image's structure. The ISP helps the model to better distinguish between foreground and background elements, improving segmentation accuracy in challenging real-world scenarios.
To handle images of varying sizes, we implement a padding strategy during batching. Each image is padded to match the size of the largest image in the batch, ensuring that all images and their corresponding labels are consistently aligned. This approach allows the model to process batches efficiently without requiring resizing or cropping, which could potentially lead to information loss or distortion in the data. By ensuring consistency in image dimensions across the batch, we improve the efficiency and accuracy of the training process, while also maintaining the integrity of the input data.

\textbf{Training Strategy.} We keep the pretrained SAM frozen throughout training to preserve its rich visual representations learned from large-scale datasets. Only the newly introduced parameters in \ourmodel, designed to enhance segmentation under complex scenarios, are optimized. The model is trained for 12 epochs with an initial learning rate of 0.001, which is decayed after epoch 10 to enhance training stability and prevent overfitting. All experiments are conducted on an NVIDIA RTX A6000 GPU, using the maximum feasible batch size for efficient GPU utilization and accelerated convergence. This setup ensures both training efficiency and competitive performance across standard benchmarks. For a controlled architectural comparison, every method in \Cref{tab: compare_sam} is retrained on this identical training set under the same schedule, resolution, and box prompts---SAM-based methods from their corresponding SAM initialization---so the reported differences isolate the effect of architectural design rather than training data or supervision.

\textbf{Inference Protocol.} During inference, the boundary loss prediction branch is disabled to simplify the evaluation process and focus on final mask generation. SAM-generated masks are fused at the logit level with predictions from \ourmodel, producing intermediate results at a resolution of $256 \times 256$. These are subsequently upsampled to $1024 \times 1024$ using bilinear interpolation. This hierarchical refinement strategy preserves structural integrity while maintaining high segmentation accuracy in the final output.

\textbf{Evaluation Metrics.} To accurately quantify improvements in mask quality, instead of only employing the standard mask AP or mask mIoU, we also adopt the boundary metric mBIoU to measure contour alignment.
Formally, let $N_c$ be the number of classes, and $P_c, G_c$ denote the predicted and ground truth masks for class $c$.
The mIoU assesses region-based accuracy:
\begin{equation}
    \text{mIoU} = \frac{1}{N_c} \sum_{c=1}^{N_c} \frac{|P_c \cap G_c|}{|P_c \cup G_c|}.
\end{equation}
Complementarily, mBIoU computes the IoU strictly within a distance $d$ from the contours (denoted as regions $P_c^d$ and $G_c^d$):
\begin{equation}
    \text{mBIoU} = \frac{1}{N_c} \sum_{c=1}^{N_c} \frac{|P_c^d \cap G_c^d|}{|P_c^d \cup G_c^d|}.
\end{equation}

\subsection{Implementation Details}
All experiments are implemented using PyTorch~\cite{paszke2019pytorch} (v1.13.1+cu116) to ensure reproducibility in fine-grained, real-world segmentation research. To maintain consistency and facilitate benchmark comparisons, we adopt the SAM~\cite{kirillov2023segment} inference pipeline and incorporate the HQ-output token from HQ-SAM~\cite{ke2023segment} for efficient mask prediction. This design enables our model to leverage the flexibility and precision of the SAM framework while enhancing its performance to handle more complex segmentation tasks. 

For bounding box prompt evaluations, identical bounding boxes are provided to all SAM variants with single-mask output mode to ensure fair comparisons.


\subsection{Comparisons with the State-of-the-art Methods}
\subsubsection{Quantity Analysis}

\begin{table*}[t]
\centering
\caption{\ourmodel\ vs. SAM-Related Models on our proposed \ourdataset\ and four extremely
fine-grained benchmarks, where the light purple represents our model.}
\vspace{0.05in}
\setlength{\tabcolsep}{2mm}
\resizebox{1.0\linewidth}{!}{
\begin{tabular}{l|cccccccccccc|cc}
\toprule
\multicolumn{1}{l}{\multirow{2}{*}{Model}} &
  \multicolumn{2}{c}{\ourdataset} &  \multicolumn{2}{c}{DIS} & \multicolumn{2}{c}{COIFT}& \multicolumn{2}{c}{HRSOD} & \multicolumn{2}{c}{ThinObject} & \multicolumn{2}{c}{Average}& \multicolumn{2}{c}{Params} \\
\multicolumn{1}{c}{} &
  mIoU&
  mBIoU &  mIoU& mBIoU& mIoU&mBIoU& mIoU& mBIoU& mIoU&mBIoU & mIoU&mBIoU& Total (M)&Learnable (M) \\ \midrule
 VPD~\cite{zhao2023unleashing}& 77.2& 61.5& 63.2& 67.8& 86.2&80.6& 86.3& 76.9& 60.2&54.3& 74.6&68.2& 901.8& 867.6\\
 MaskFormer~\cite{cheng2021per}& 78.6& 62.8& 61.7& 66.5& 87.4& 81.2& 88.8& 77.2& 62.1& 56.9& 75.7& 68.9& 16.5&16.5\\
 SAM-B ~\cite{kirillov2023segment}& 79.9& 63.9& 53.6& 45.1& 87.9& 81.3& 86.1& 76.9& 54.1& 44.8& 72.3& 62.4& 93.7&93.7\\
 HQ-SAM-B~\cite{ke2023segment} & \textbf{81.4}& 67.1 & 76.3&68.3& 93.1&88.0& 91.2& \textbf{84.1}& 84.5&73.4 & 85.3&76.2& 94.8& 1.07 \\
 \rowcolor[rgb]{ .894,  .875,  .925}  \ourmodel-B~(Ours) & 81.1& 67.1 & \textbf{77.7}&\textbf{70.4}& \textbf{93.7}&\textbf{88.7}& \textbf{91.4}& 83.9& \textbf{87.4}&\textbf{77.6} & \textbf{86.3}&\textbf{77.5}& 94.8& 1.11 \\ \midrule
SAM-L~\cite{kirillov2023segment} &
  83.3&
  68.7 &  62.0&52.8& 92.1&86.5& 90.2& 83.1& 73.6&61.8 & 80.2&70.6& 312.3& 312.3 \\
HQ-SAM-L~\cite{ke2023segment} &
  83.9&
  70.2 &  78.6&70.4& \textbf{94.8}&90.1& \textbf{93.6}& 86.9& 89.5&79.9 & 88.0&79.5& 313.7& 1.33 \\
\rowcolor[rgb]{ .894,  .875,  .925}   \ourmodel-L~(Ours) & \textbf{84.6}& \textbf{70.9}& \textbf{80.2}&\textbf{73.7}& 94.6&\textbf{90.2}& 93.0& 86.9& \textbf{91.3}&83.4 & \textbf{88.7}&\textbf{81.0}& 313.7& 1.37 \\ \midrule
SAM-H~\cite{kirillov2023segment} &
  83.5&
  69.3 &  57.1&49.3& 90.8&85.6& 87.0& 80.1& 68.4&58.3 & 77.4&68.5& 641.1& 641.1 \\
HQ-SAM-H ~\cite{ke2023segment} &
  83.8&
  69.2 &  79.2&71.4& 95.0&90.3& 92.1& 84.7& 89.2&79.6 & 87.9&79.0& 642.7& 1.60 \\
\rowcolor[rgb]{ .894,  .875,  .925}  \ourmodel-H~(Ours) &
  \textbf{84.4}&
  \textbf{70.6}  &  \textbf{81.4}& \textbf{75.0}& 95.0&\textbf{90.6}& \textbf{92.6}& \textbf{86.7}& \textbf{91.8}&\textbf{84.1} & \textbf{89.0}&\textbf{81.4}& 642.7& 1.63 \\ \bottomrule
\end{tabular}
}
\label{tab: compare_sam}
\end{table*}

As demonstrated in~\Cref{tab: compare_sam}, we compared \ourmodel\ with various types of models across five datasets, emphasizing its notable advantages in several key aspects:

\textbf{In terms of Accuracy:} \ourmodel\ outperforms others on most of the datasets, with an 89.0\% mean IoU on average, improving by 1.1\% compared to the HQ-SAM. Particularly, on the fine-grained ThinObject dataset, the \textbf{B}ase, \textbf{L}arge, and \textbf{H}uge versions of \ourmodel\ significantly outperform other methods. Meanwhile, \ourmodel\ demonstrates the strongest robustness on the \ourdataset\ dataset. 

\textbf{In terms of Efficiency:} \ourmodel's parameter scale is comparable to the HQ-SAM series, achieves significantly superior performance, and possesses considerably fewer parameters than VPD~\cite{zhao2023unleashing}. For instance, \ourmodel-L on the DIS dataset increases mIoU from 78.6 to 80.2 with a minimal increase of only $\sim$0.035M parameters, indicating its enhanced learning capacity compared to models relying on isolated spatial-domain learning.

\Cref{tab: efficiency} reports the full cost profile measured on a single RTX A6000 at $1024\times1024$ input. The three modules add $0.035$\,M trainable parameters and about $2$ GFLOPs, which is $0.07\%$ of the ViT-H budget, leave peak memory unchanged and reduce throughput by $1$ to $3\%$. Measured in isolation, the forward and inverse FFT together cost $0.34$ GFLOPs and under $0.5$\,ms, and the $256\times256$ feature map yields a $256\times129$ half spectrum, so the non-square shape is produced by the real FFT itself and requires no padding.

\begin{table}[!t]
\centering
\caption{Efficiency profile measured on a single NVIDIA RTX A6000 at $1024\times1024$ input.}
\label{tab: efficiency}
\resizebox{\linewidth}{!}{
\begin{tabular}{llrrrr}
\toprule
Backbone & Method & Params (M) & GFLOPs & Peak memory & FPS \\
\midrule
ViT-B & HQ-SAM & 94.81 & 495 & 0.5\,GB & 8.60 \\
ViT-B & \ourmodel & 94.84 & 497 & 0.5\,GB & 8.41 \\
ViT-H & HQ-SAM & 642.69 & 2993 & 2.8\,GB & 2.05 \\
ViT-H & \ourmodel & 642.72 & 2995 & 2.8\,GB & 2.03 \\
\bottomrule
\end{tabular}}
\end{table}

\begin{table}[!t]
\centering
\caption{Zero-Shot Instance Segmentation Results on COCO benchmark prompted with Grounding DINO~\cite{liu2024grounding} Boxes.}
\vspace{0.05in}
\begin{tabular}{l|cccc|c}
\toprule
\multirow{2}{*}{Model}& \multicolumn{4}{c}{GroundingDINO}                   & Params\\
                    & $AP$&$AP^{\text{S}}$&$AP^{\text{M}}$&$AP^{\text{S}}$& Learnable (M)\\
\midrule
FastSAM~\cite{zhao2023fast}            & 35.7         & 22.1& 42.1&48.2& 68.0\\
MobileSAM~\cite{zhang2023faster}         & 41.4         & 25.7& 45.6&60.5& 21.9              \\
SlimSAM-77~\cite{chen2024slimsam}         & 42.1         & 27.1& 46.0&59.6& 9.8              \\
TinySAM~\cite{shu2025tinysam}          & 42.4         & 27.4& 46.5&60.6& 10.1              \\
EdgeSAM~\cite{zhou2023edgesam}       & 42.5         & 27.4& 46.9&59.5& 9.6              \\
SlimSAM-50~\cite{chen2024slimsam}         & 43.5         & 28.7& 47.8&60.8& 28.0             \\
RepViT-SAM~\cite{wang2023repvit}            & 43.8         & 28.3& 48.2&61.6& 27.2             \\
EfficientSAM-L0~\cite{xiong2024efficientsam}   & 46.0         & 29.2& 50.3&65.7& 34.8             \\
SAM2.1-L~\cite{ravi2025sam2}& 45.6& 28.3& 50.7& 64.2& 224.3\\
SAM-H~\cite{kirillov2023segment}& 44.9& \textbf{29.6}& 49.4& 62.3& 641.1\\
HQ-SAM~\cite{ke2023segment}& 45.9& 28.8& 50.1&64.0& \textbf{1.60}\\ \midrule
\rowcolor[rgb]{ .894,  .875,  .925} \ourmodel (Ours)& \textbf{46.2}& 28.9& 50.7&\textbf{64.5}& 1.63\\ 
\bottomrule
\end{tabular}
\label{tab: zeroshot-coco}
\end{table}

\textbf{Results on COCO:} As shown in~\Cref{tab: zeroshot-coco}, \ourmodel\ outperforms FastSAM by 10.5\%, achieving the best scores across various object sizes, with only $\sim$1.63M learnable parameters (cf.\ \Cref{tab: compare_sam}), approximately 17\% of EdgeSAM, while leading EdgeSAM by 3.7\% in average precision (AP). At the same time, \ourmodel\ surpasses the state-of-the-art model SAM2.1 by 0.6\%, demonstrating strong capabilities.


\begin{table*}[!t]
\centering
\caption{Zero-Shot Segmentation Performance on the SegInW Benchmark (25 Datasets) using \ourmodel-L.}
    \resizebox{1\linewidth}{!}{
\begin{tabular}{c|c|c|c|c|c|c|c|c|c} \toprule
Models & ViT & GroundingDINO & Text-Encoder & Mean AP & Airplane-Parts & Bottles & Brain-Tumor & Chicken & Cows \\ 
\midrule
SAM~\cite{kirillov2023segment}& h & swin-b & bert & 48.7 & 37.2 & 65.4 & 11.9 & 84.5 & 47.5 \\
\rowcolor[rgb]{ 0.85, 0.92, 1 }HQ-SAM~\cite{ke2023segment}& h & swin-b & bert & 49.6 & 37.6 & 66.3 & 12.0& 84.5 & 47.8 \\
SAM2~\cite{ravi2025sam2}& l & swin-b & bert & 49.5 & 38.3 & 67.1 & 12.1 & 80.7 & 52.8 \\
\rowcolor[rgb]{ .894,  .875,  .925} \ourmodel & h & swin-b & bert & 49.7 & 36.4 & 66.0& 11.8 & 85.6 & 47.7 \\ \midrule
Electric-Shaver & Elephants & Fruits & Garbage & Ginger-Garlic & Hand-Metal & Hand & House-Parts & HouseHold-Items & Nutterfly-Squireel \\ \midrule
71.7& 77.9 & 82.3 & 24.0& 45.8 & 81.2 & 70.0& 8.4 & 60.1 & 71.3 \\
\rowcolor[rgb]{ 0.85, 0.92, 1 }72.1& 77.5 & 82.3 & 25.0& 45.6 & 81.2 & 74.8 & 8.5& 60.1 & 77.1 \\
72.0& 78.2 & 83.3 & 26.0& 45.7 & 73.7 & 77.6 & 8.6& 60.1 & 84.1 \\
\rowcolor[rgb]{ .894,  .875,  .925} 72.4& 77.6 & 82.0& 24.3 & 46.1 & 77.0& 81.2 & 8.6 & 60.1 & 77.6 \\ \midrule
Phones & Poles & Puppies & Rail & Salmon-Fillet & Strawberry & Tablets & Toolkits & Trash & Watermelon \\ \midrule
35.4& 23.3 & 50.1 & 8.7 & 32.9 & 83.5 & 29.8 & 20.8 & 30.0& 64.2 \\
\rowcolor[rgb]{ 0.85, 0.92, 1 }35.3& 20.1 & 50.1 & 7.7 & 42.2 & 85.6 & 29.7 & 21.8 & 30.0& 65.6 \\
34.6& 28.8 & 48.9 & 14.3 & 24.2 & 83.7 & 29.1 & 20.1 & 28.4 & 66.0\\
\rowcolor[rgb]{ .894,  .875,  .925} 34.5& 21.1 & 50.1 & 8.4 & 42.3 & 85.8 & 29.4 & 21.2 & 29.8 & 65.2\\
 \bottomrule  
\end{tabular}}
\label{tab: SeginW 25 categories}
\end{table*}

\textbf{Results on SegInW:} As presented in~\Cref{tab: SeginW 25 categories}, we evaluate the zero-shot segmentation performance on the SegInW benchmark, which comprises 25 distinct unstructured segmentation tasks. \ourmodel\ achieves an overall Mean Average Precision (Mean AP) of 49.7\%, surpassing the SAM (48.7\%) by 1.0\%. This result underscores our model's exceptional ability to generalize across diverse real-world scenarios. Compared to state-of-the-art methods, \ourmodel\ demonstrates varying degrees of improvement, particularly distinguishing itself in the segmentation of everyday objects. For instance, it outperforms competing models in categories such as Electric-Shaver, Hand, and Strawberry, validating its robustness in processing complex visual data.

\subsubsection{Quality Analysis}

\begin{figure*}[!h]
  \centering
  \includegraphics[width=1\textwidth]{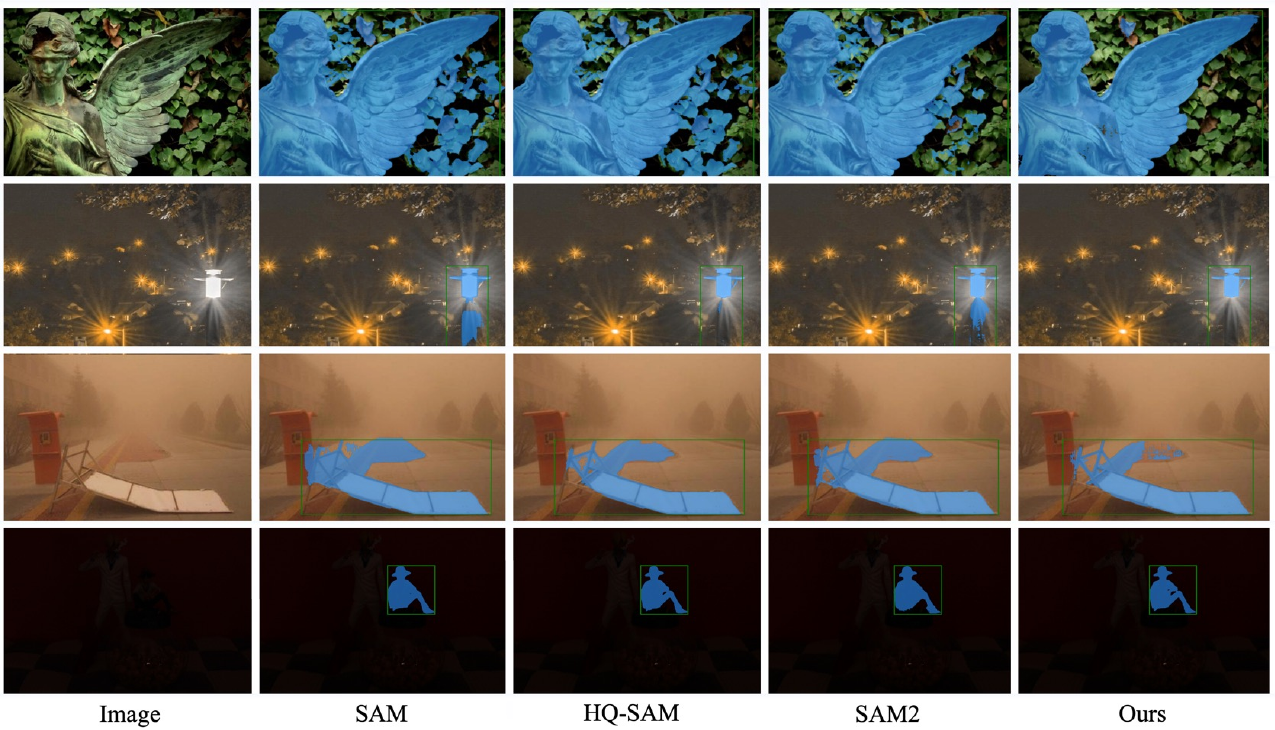}
  \caption{The segmentation comparison of SAM-related models across multiple datasets. The corner inset highlights segmentation details.}
  \label{fig: show_seg}
\end{figure*}

As illustrated in \Cref{fig: show_seg}, we present a qualitative comparison of segmentation results across multiple datasets. While state-of-the-art models like SAM, HQ-SAM, and SAM2 often struggle under complex environmental conditions, \ourmodel~demonstrates superior robustness and semantic understanding. First, in scenarios with complex background clutter (Row 1), competitors tend to over-segment, confusing the statue's wings with the surrounding foliage. In contrast, our method precisely delineates the object boundaries, effectively suppressing irrelevant background textures. Second, under adverse lighting conditions with strong optical glare (Row 2), spatial-only models are distracted by the light beams, resulting in erroneous masks that include the glare itself. Our model, however, correctly identifies the physical structure of the street lamp, ignoring the visual artifacts. Third, in severe weather scenarios like the sandstorm (Row 3), where visibility is degraded and contrast is low, competitors produce fragmented and incomplete masks. Our approach successfully recovers the holistic structure of the object, verifying its capability to handle extreme visual degradation. Finally, in extremely low-light environments with negligible foreground-background contrast (Row 4), baseline models fail to discern the target boundaries, resulting in vague or missed detections. Our approach accurately captures the fine-grained silhouette of the target, demonstrating remarkable robustness to severe illumination changes. Overall, these results confirm that \ourmodel~not only preserves internal structural integrity but also minimizes false segmentation in challenging, unseen environments.

\subsection{Ablation Study}

\begin{table}[!t]
\centering
\caption{Component effectiveness on widely-recognized dataset DUTS (ViT-B; purple: \ourmodel, blue: baseline).}
\vspace{0.05in}
\setlength{\tabcolsep}{2mm}
\resizebox{0.9\linewidth}{!}{
\begin{tabular}{l|cccc|cc}
\toprule
Model&SPM  &DDAM&EEM& Mask Feature& mIoU& mBIoU\\ 
\midrule
\rowcolor[rgb]{ 0.85, 0.92, 1 }HQ-SAM~\cite{ke2023segment}&           &&           &            \checkmark& 86.53 (0.06)& 73.08 (0.07)\\
 & && \checkmark&  \checkmark  & 87.84 (0.04)&75.15 (0.04)\\
 &  & \checkmark& & \checkmark& 88.04 (0.01)&75.69 (0.04)\\
 \rowcolor[rgb]{ .741,  .937,  .741} Without FFT& & \checkmark& & \checkmark& 86.96 (0.06)&73.21 (0.08)\\
 & \checkmark& & & \checkmark& 87.91 (0.10)&75.18 (0.08)\\
 \rowcolor[rgb]{ .949,  .996,  .627}Without ISP& \checkmark& & & \checkmark& 87.16 (0.03)&75.06 (0.03)\\
 & \checkmark& \checkmark& & \checkmark& 88.22 (0.13)&75.80 (0.13)\\
 & \checkmark& & \checkmark& \checkmark& 88.12 (0.10)&75.88 (0.08)\\
           &            &\checkmark&           \checkmark&           \checkmark  & 89.26 (0.07)& 75.72 (0.06)\\\rowcolor[rgb]{ .894,  .875,  .925}{\ourmodel}& \checkmark &\checkmark& \checkmark& \checkmark  & 89.61 (0.03)& 75.78 (0.04)\\ \bottomrule
\end{tabular}}
\label{tab: abla-self}
\end{table}


To verify the effectiveness of each component, we conducted a comprehensive ablation study on the DUTS dataset. To ensure statistical reliability, we report the mean IoU and standard deviation across multiple fixed-seed runs in \Cref{tab: abla-self}. 

\textbf{Analysis of SPM:} SPM significantly enhances performance by mitigating global noise and providing structural priors to EEM. Without ISP, SPM contributes only a 0.63\% mIoU gain; with ISP, the gain increases to 1.38\%. Together with EEM, SPM achieves a 1.59\% improvement. This progressive gain substantiates the synergistic effect between global context modeling and local structural refinement, confirming that SPM functions not merely as a feature extractor, but as a critical structural hub that aligns multi-scale representations.

\textbf{Analysis of DDAM:} DDAM offers a 1.51\% mIoU and 2.61\% mBIoU gain, validating the dual-domain decoding strategy. LSEB alone yields 86.96 mIoU, while the addition of FDFB (FFT-based) boosts performance to 88.04 mIoU. This significant improvement highlights the limitations of spatial-only processing and demonstrates that integrating frequency-domain constraints effectively captures global noise patterns that are indistinguishable in the spatial domain.

\textbf{Analysis of EEM:}
EEM improves boundary perception, leading to $\sim$1.5\% gains in both metrics. This result indicates that explicitly targeting boundary artifacts is essential for segmentation quality, as it resolves ambiguities in transition regions where standard pixel-wise classification typically fails.

\begin{table}[!t]
    \centering
    \caption{Comparison of Different Convolution kernel sizes of SPM and Compress Ratio of $CRA_m$ on \ourdataset\ benchmark with ViT-B backbone.}
    \resizebox{0.9\linewidth}{!}{
    \begin{tabular}{c}
    \begin{minipage}{0.48\linewidth}
        \centering
        \textbf{(a) Convolution kernel variants} \\[1ex]
        \begin{tabular}{cc|cc}
            \toprule
            \multicolumn{2}{c}{Filter} & mIoU & mBIoU\\ 
            \midrule
            $3\times3$ & $5\times5$ & 84.90 & 71.09 \\
            $3\times3$ & $7\times7$ & 84.95 & 71.25 \\
            $3\times3$ & $9\times9$ & 85.02 & 71.27 \\
            $5\times5$ & $7\times7$ & 85.08 & 71.15 \\
            $5\times5$ & $9\times9$ & 84.98 & 71.15 \\
            \rowcolor[rgb]{ .894,  .875,  .925} $7\times7$ & $9\times9$ & 85.09 & 71.36 \\
            \bottomrule
        \end{tabular}
    \end{minipage}
    \hspace{0.04\linewidth}
    \begin{minipage}{0.48\linewidth}
        \centering
        \textbf{(b) Compress Ratio} \\[1ex]
        \begin{tabular}{c|cc}
            \toprule
            Compress Ratio & mIoU & mBIoU\\ 
            \midrule
            1  & 84.99 & 71.09 \\
            \rowcolor[rgb]{ .894,  .875,  .925} 2 & 85.12 & 71.26 \\
            \rowcolor{white}
            4  & 84.72 & 70.79 \\
            8  & 84.84 & 71.01 \\
            16 & 84.89 & 71.20 \\
            32 & 85.04 & 71.24 \\
            \bottomrule
        \end{tabular}
    \end{minipage}
    \end{tabular}
    }
    \label{tab: abla-GroupConv&CompressRatio}
\end{table}

\textbf{Analysis of Different Convolution Kernels of SPM:}
As shown in \cref{tab: abla-GroupConv&CompressRatio} (a), $7\times7$ and $9\times9$ group convolutions outperform smaller kernels by $\sim$0.2\%. This finding underscores the importance of larger receptive fields in capturing long-range dependencies, which are requisite for distinguishing foreground objects from complex backgrounds.

\textbf{Analysis of Compression Ratio of $CRA_m$ in the DDAM:}
\cref{tab: abla-GroupConv&CompressRatio} (b) shows that compressing Q and K vectors improves global representation. A compression ratio of 2 (channels reduced to 16) improves mIoU by 0.13\% over the full configuration. This suggests that a moderate reduction in channel redundancy acts as a regularizer, preventing the model from overfitting to high-frequency noise while maintaining computational efficiency.

\begin{figure}[!h]
  \centering
  \includegraphics[scale=0.3]{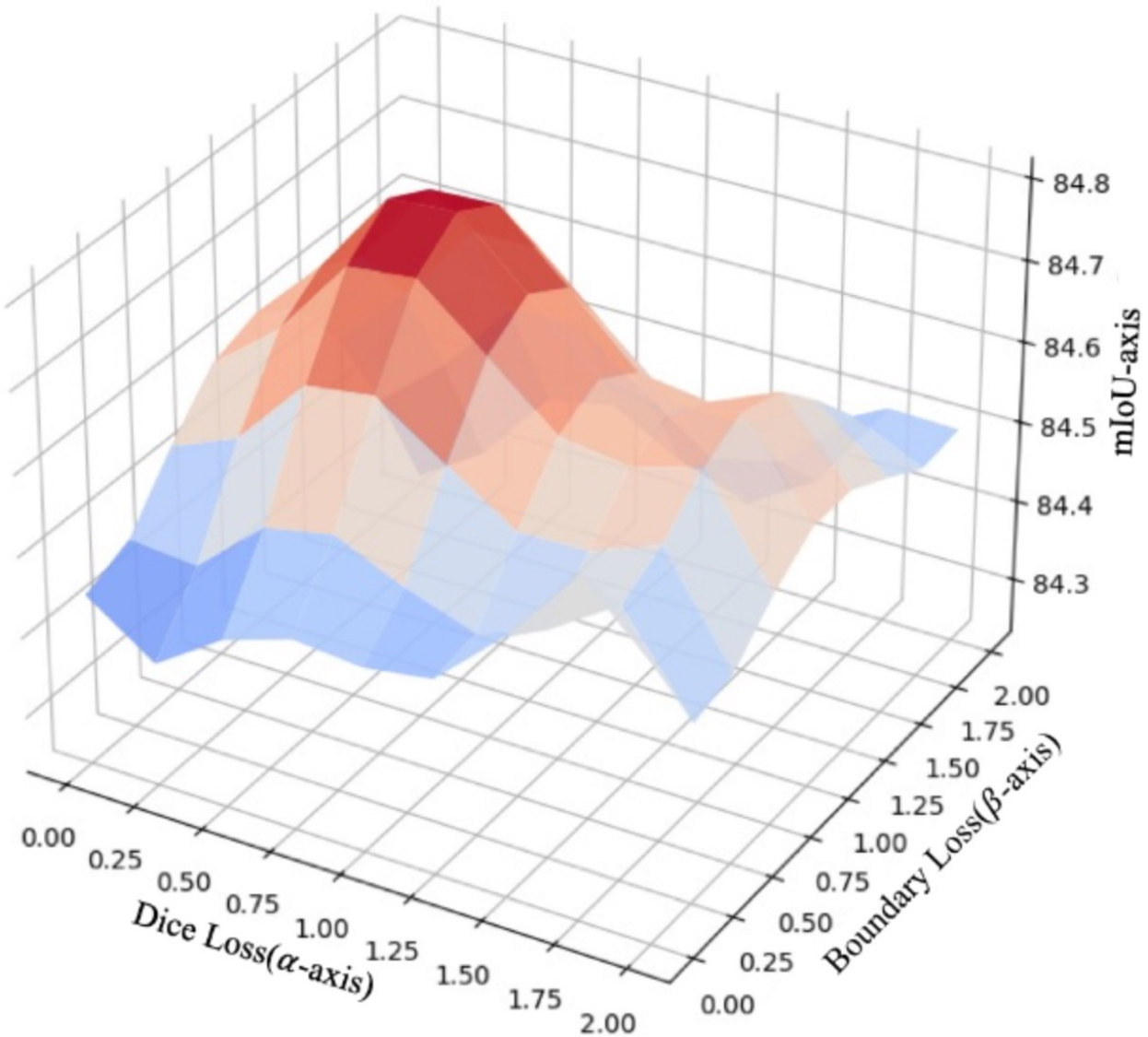}
  \caption{Visualization of Loss with mIoU-axis, $\mathcal{L}_{\text{Dice}}$($\alpha$-axis) and $\mathcal{L}_{\text{Boundary}}$($\beta$-axis) on the \ourdataset\ dataset processed by \ourmodel\ with ViT-B.}
  \label{fig: abla-loss-grid-search}
\end{figure}

\textbf{Boundary Loss and Dice Loss Analysis:} Grid search (\Cref{fig: abla-loss-grid-search}) reveals optimal performance at $\alpha:\beta = 1:0.75$, achieving 85.20 mIoU and 71.55 mBIoU. This ratio balances segmentation accuracy and boundary precision, underscoring the value of boundary-aware loss design.

\begin{table}[tbp]
  \centering
  \caption{Comparison of filter variants in the FDFB module on the HQSeg-44k benchmark using a ViT-L backbone. \textbf{Note:} Learnable parameter counts are reported as the number of parameters (not memory size in bytes).}
  \begin{tabular}{c|ccc}
    \toprule
    \textbf{Filter} & \textbf{mIoU} & \textbf{mBIoU} &\textbf{Learnable Params}\\ 
    \midrule
    SE Block & 89.96 & 81.41 &1368125\\
    ECA Block& 90.12 & 81.79 &1367875\\
    \rowcolor[rgb]{ .894,  .875,  .925}CBAM Block& 91.81 & 83.49 &1368321\\
    \bottomrule
  \end{tabular}
  \label{tab: abla-filter}
\end{table}

\textbf{Filter Selection.} We evaluated the effect of various dynamic filters on model performance, as shown in~\Cref{tab: abla-filter}, including the Base Filter, SE block~\cite{hu2018squeeze}, ECA~\cite{wang2020eca}, and CBAM~\cite{woo2018cbam}. Compared to SE, ECA reduced parameter count by 250 with a 0.16\% gain in mBIoU. CBAM, despite adding 196 parameters, achieved a notable 1.85\% improvement, reaching 91.81\% mIoU. Considering the trade-off between accuracy and efficiency, CBAM is adopted for its superior performance with minimal overhead.

The full model achieves 89.61\% mIoU and 75.78\% mBIoU, outperforming the baseline by 3.08\% and 2.7\%, respectively, highlighting the efficacy of component integration in \ourmodel.

\subsection{Network Interpretability}

\begin{figure*}[!t]
  \centering
  \includegraphics[width=1\textwidth]{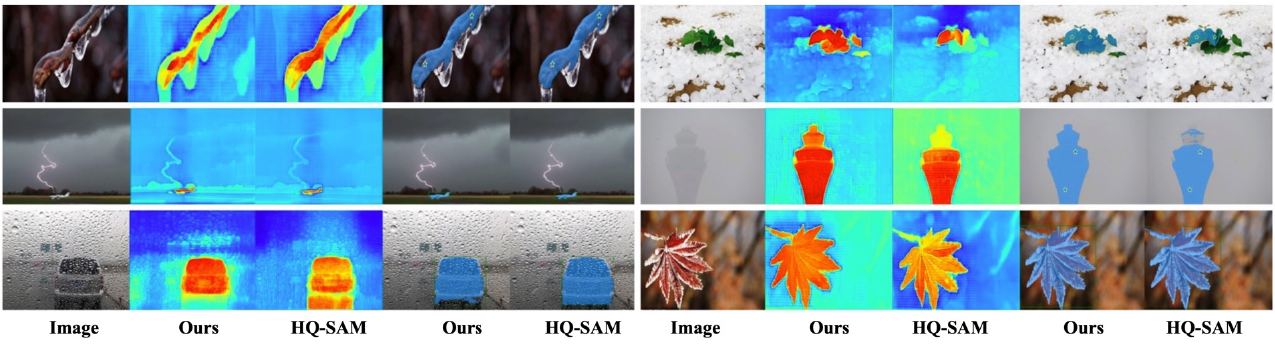}
  \caption{Visual comparison of cross-attention heatmaps between the original HQ-SAM output token and the \ourmodel\ output token in the final decoder layer. \ourmodel\ demonstrates stronger focus on object boundaries and structural regions that are misclassified by the original token.}
  \label{fig:hotmap}
\end{figure*}

As shown in \Cref{fig:hotmap}, we visualize the output tokens generated by \ourmodel\ and HQ-SAM in the token-to-image cross-attention layer of the mask decoder. The results indicate that compared to HQ-SAM, \ourmodel's attention mechanism is more concentrated on the true structural boundaries of the target objects. Specifically, in the top-left example, the spatial domain effectively integrates semantically relevant branch information, while the frequency domain---guided by our designed cross-domain solver---suppresses background interference, thereby enhancing the accurate localization of target structures.

We further probe what the trained filter does, by hooking the FDFB and recording the ratio of output to input energy in the high band. The filter is adaptive rather than fixed: it amplifies the high band by $3.47\times$ on blurred input and by $3.45\times$ on haze, where high frequencies are recoverable detail, but by only $2.89\times$ on additive noise, which is $0.88$ of its gain on clean input. The phase pathway remains close to identity, so edge positions are preserved. This is the behaviour the module is designed for, recovering structure where it is missing while refusing to amplify noise.

\section{Future Work}
Building upon the promising results of \ourmodel\ and the \ourdataset\ benchmark, our future research will focus on two strategic directions to further advance the field of robust segmentation.

\textbf{Expansion of the \ourdataset\ Ecosystem.} 
While the current iteration of \ourdataset\ covers ten challenging non-ideal scenarios, the complexity of real-world environments is virtually boundless. We intend to significantly augment the dataset by incorporating a broader spectrum of complex scenarios, including extreme weather conditions, rare visual artifacts, and highly dynamic occlusions. By expanding the semantic categories and increasing the diversity of adverse conditions, we aim to establish a more comprehensive and rigorous benchmark. This initiative is expected to facilitate the development of foundation models with superior generalization capabilities and drive community-wide progress in addressing long-tail distribution problems in segmentation tasks.

\textbf{Lightweight Architecture and Edge Deployment.} 
Although \ourmodel\ achieves state-of-the-art performance, the reliance on sophisticated dual-domain processing and structural priors entails computational costs that may hinder deployment on resource-constrained platforms. To address this, we plan to explore lightweight backbone architectures and efficient attention mechanisms that maintain high accuracy while drastically reducing memory footprint and parameter count. Furthermore, we will investigate model compression techniques, such as knowledge distillation and quantization, to optimize \ourmodel\ for real-time inference. Our ultimate goal is to bridge the gap between high-performance research models and practical applications, enabling the deployment of robust segmentation systems on mobile and edge devices.

\section{Conclusions}
In conclusion, we introduce \ourmodel, the first segmentation framework with inherent noise-awareness, incorporating structural priors and dual-domain contextual understanding to effectively address the challenges of real-world segmentation. \ourmodel\ structure prior, dual-domain awareness, and edge estimation modules work synergistically to recover fine details, disentangle noise, and enhance boundary precision. \ourmodel\ outperforms state-of-the-art methods in fine-grained tasks and demonstrates  strong robustness across diverse scenarios. Evaluations of seven benchmarks further validate its effectiveness. Additionally, the introduction of \ourdataset, with 10 categories, establishes a new benchmark for segmentation research. This work highlights the potential of prior knowledge and cross-domain enhancements, offering valuable insights for developing computation-efficient models.

\section*{Acknowledgments}

\bibliographystyle{IEEEtran}
\bibliography{FSA_refs}

\begin{thebibliography}{10}
\providecommand{\url}[1]{#1}
\csname url@samestyle\endcsname
\providecommand{\newblock}{\relax}
\providecommand{\bibinfo}[2]{#2}
\providecommand{\BIBentrySTDinterwordspacing}{\spaceskip=0pt\relax}
\providecommand{\BIBentryALTinterwordstretchfactor}{4}
\providecommand{\BIBentryALTinterwordspacing}{\spaceskip=\fontdimen2\font plus
\BIBentryALTinterwordstretchfactor\fontdimen3\font minus
  \fontdimen4\font\relax}
\providecommand{\BIBforeignlanguage}[2]{{%
\expandafter\ifx\csname l@#1\endcsname\relax
\typeout{** WARNING: IEEEtran.bst: No hyphenation pattern has been}%
\typeout{** loaded for the language `#1'. Using the pattern for}%
\typeout{** the default language instead.}%
\else
\language=\csname l@#1\endcsname
\fi
#2}}
\providecommand{\BIBdecl}{\relax}
\BIBdecl

\bibitem{lu2024lm}
Z.~Lu, C.~She, W.~Wang, and Q.~Huang, ``Lm-net: A light-weight and multi-scale
  network for medical image segmentation,'' \emph{Computers in Biology and
  Medicine}, vol. 168, p. 107717, 2024.

\bibitem{muralidhara2025domain}
S.~Muralidhara, R.~Schuster, and D.~Stricker, ``Domain-incremental semantic
  segmentation for autonomous driving under adverse driving conditions,''
  \emph{arXiv preprint arXiv:2501.05246}, 2025.

\bibitem{hurtado2022semantic}
J.~V. Hurtado and A.~Valada, ``Semantic scene segmentation for robotics,'' in
  \emph{Deep learning for robot perception and cognition}.\hskip 1em plus 0.5em
  minus 0.4em\relax Elsevier, 2022, pp. 279--311.

\bibitem{ronneberger2015u}
O.~Ronneberger, P.~Fischer, and T.~Brox, ``U-net: Convolutional networks for
  biomedical image segmentation,'' in \emph{International Conference on Medical
  image computing and computer-assisted intervention}.\hskip 1em plus 0.5em
  minus 0.4em\relax Springer, 2015, pp. 234--241.

\bibitem{elgamily2025novel}
K.~M. Elgamily, M.~Mohamed, A.~M. Abou-Taleb, and M.~M. Ata, ``A novel w13 deep
  cnn structure for improved semantic segmentation of multiple objects in
  remote sensing imagery,'' \emph{Neural Computing and Applications}, pp.
  1--31, 2025.

\bibitem{cao2024medsegmamba}
A.~Cao, Z.~Li, J.~Jomsky, A.~F. Laine, and J.~Guo, ``Medsegmamba: 3d cnn-mamba
  hybrid architecture for brain segmentation,'' \emph{arXiv preprint
  arXiv:2409.08307}, 2024.

\bibitem{xie2021segformer}
E.~Xie, W.~Wang, Z.~Yu, A.~Anandkumar, J.~M. Alvarez, and P.~Luo, ``Segformer:
  Simple and efficient design for semantic segmentation with transformers,'' in
  \emph{Advances in Neural Information Processing Systems}, vol.~34, 2021, pp.
  12\,077--12\,090.

\bibitem{zheng2021rethinking}
S.~Zheng, J.~Lu, H.~Zhao, X.~Zhu, Z.~Luo, Y.~Wang, Y.~Fu, J.~Feng, T.~Xiang,
  P.~H. Torr \emph{et~al.}, ``Rethinking semantic segmentation from a
  sequence-to-sequence perspective with transformers,'' in \emph{Proceedings of
  the IEEE/CVF conference on computer vision and pattern recognition}, 2021,
  pp. 6881--6890.

\bibitem{chai2025transdeep}
T.~Chai, Z.~Xiao, X.~Shen, Q.~Liu, N.~Li, T.~Guan, and J.~Tian, ``Transdeep:
  Transformer-integrated deeplabv3+ for image semantic segmentation,''
  \emph{IEEE Access}, 2025.

\bibitem{lin2014microsoft}
T.-Y. Lin, M.~Maire, S.~Belongie, J.~Hays, P.~Perona, D.~Ramanan,
  P.~Doll{\'a}r, and C.~L. Zitnick, ``Microsoft coco: Common objects in
  context,'' in \emph{European conference on computer vision}.\hskip 1em plus
  0.5em minus 0.4em\relax Springer, 2014, pp. 740--755.

\bibitem{Zhou2023Image}
R.~Zhou, L.~Zheng, C.~Ren, and S.~Wang, ``Image segmentation algorithm in
  complex environment based on improved solov2,'' in \emph{2023 12th
  International Conference of Information and Communication Technology
  (ICTech)}.\hskip 1em plus 0.5em minus 0.4em\relax IEEE, 2023, pp. 581--585.

\bibitem{drayer2016object}
B.~Drayer and T.~Brox, ``Object detection, tracking, and motion segmentation
  for object-level video segmentation,'' \emph{arXiv preprint
  arXiv:1608.03066}, 2016.

\bibitem{wang2022deep}
Z.~Wang, Y.~Zhang, K.~M. Mosalam, Y.~Gao, and S.-L. Huang, ``Deep semantic
  segmentation for visual understanding on construction sites,''
  \emph{Computer-Aided Civil and Infrastructure Engineering}, vol.~37, no.~2,
  pp. 145--162, 2022.

\bibitem{li2024saliency}
Y.~Li, W.~Gao, G.~Li, and S.~Ma, ``Saliency segmentation oriented deep image
  compression with novel bit allocation,'' \emph{IEEE Transactions on Image
  Processing}, 2024.

\bibitem{jain2023oneformer}
J.~Jain, J.~Li, M.~T. Chiu, A.~Hassani, N.~Orlov, and H.~Shi, ``Oneformer: One
  transformer to rule universal image segmentation,'' in \emph{Proceedings of
  the IEEE/CVF Conference on Computer Vision and Pattern Recognition (CVPR)},
  June 2023, pp. 2989--2998.

\bibitem{yan2023universal}
B.~Yan, Y.~Jiang, J.~Wu, D.~Wang, P.~Luo, Z.~Yuan, and H.~Lu, ``Universal
  instance perception as object discovery and retrieval,'' in \emph{Proceedings
  of the IEEE/CVF Conference on Computer Vision and Pattern Recognition
  (CVPR)}, June 2023, pp. 15\,325--15\,336.

\bibitem{luddecke2022image}
T.~L\"uddecke and A.~Ecker, ``Image segmentation using text and image
  prompts,'' in \emph{Proceedings of the IEEE/CVF Conference on Computer Vision
  and Pattern Recognition (CVPR)}, June 2022, pp. 7086--7096.

\bibitem{kirillov2023segment}
A.~Kirillov, E.~Mintun, N.~Ravi, H.~Mao, C.~Rolland, L.~Gustafson, T.~Xiao,
  S.~Whitehead, A.~C. Berg, W.-Y. Lo, P.~Dollar, and R.~Girshick, ``Segment
  anything,'' in \emph{Proceedings of the IEEE/CVF International Conference on
  Computer Vision (ICCV)}, October 2023, pp. 4015--4026.

\bibitem{zhang2023faster}
\BIBentryALTinterwordspacing
C.~Zhang, D.~Han, Y.~Qiao, J.~U. Kim, S.-H. Bae, S.~Lee, and C.~S. Hong,
  ``Faster segment anything: Towards lightweight sam for mobile applications,''
  2023. [Online]. Available: \url{https://arxiv.org/abs/2306.14289}
\BIBentrySTDinterwordspacing

\bibitem{chen2024slimsam}
\BIBentryALTinterwordspacing
Z.~Chen, G.~Fang, X.~Ma, and X.~Wang, ``Slimsam: 0.1\% data makes segment
  anything slim,'' in \emph{Advances in Neural Information Processing Systems},
  A.~Globerson, L.~Mackey, D.~Belgrave, A.~Fan, U.~Paquet, J.~Tomczak, and
  C.~Zhang, Eds., vol.~37.\hskip 1em plus 0.5em minus 0.4em\relax Curran
  Associates, Inc., 2024, pp. 39\,434--39\,461. [Online]. Available:
  \url{https://proceedings.neurips.cc/paper_files/paper/2024/file/45a7ca247462d9e465ee88c8a302ca70-Paper-Conference.pdf}
\BIBentrySTDinterwordspacing

\bibitem{chen2019visual}
Z.~Chen and H.~Zhu, ``Visual quality evaluation for semantic segmentation:
  subjective assessment database and objective assessment measure,'' \emph{IEEE
  Transactions on Image Processing}, vol.~28, no.~12, pp. 5785--5796, 2019.

\bibitem{zhang2022enhanced}
K.~Zhang, D.~Li, W.~Luo, W.~Ren, and W.~Liu, ``Enhanced spatio-temporal
  interaction learning for video deraining: Faster and better,'' \emph{IEEE
  Transactions on Pattern Analysis and Machine Intelligence}, vol.~45, no.~1,
  pp. 1287--1293, 2022.

\bibitem{zhang2021deep}
K.~Zhang, R.~Li, Y.~Yu, W.~Luo, and C.~Li, ``Deep dense multi-scale network for
  snow removal using semantic and depth priors,'' \emph{IEEE Transactions on
  Image Processing}, vol.~30, pp. 7419--7431, 2021.

\bibitem{jin2025mb}
Z.~Jin, Y.~Qiu, K.~Zhang, H.~Li, and W.~Luo, ``Mb-taylorformer v2: Improved
  multi-branch linear transformer expanded by taylor formula for image
  restoration,'' \emph{IEEE Transactions on Pattern Analysis and Machine
  Intelligence}, vol.~47, no.~7, pp. 5990--6005, 2025.

\bibitem{zhao2026echosr}
H.~Zhao, B.~Wang, S.~Zhao, T.~Wang, K.~Zhang, and W.~Lu, ``Echosr: Efficient
  context harnessing for lightweight image super-resolution,''
  \emph{Information Fusion}, p. 104471, 2026.

\bibitem{liu2025condition}
Y.~Liu, X.~Dong, Y.~Lin, M.~Ye, K.~Zhang, and B.~Du, ``Condition-guided
  diffusion for multi-modal pedestrian trajectory prediction incorporating
  intention and interaction priors,'' \emph{IEEE Transactions on Pattern
  Analysis and Machine Intelligence}, 2025.

\bibitem{xiong2024efficientsam}
Y.~Xiong, B.~Varadarajan, L.~Wu, X.~Xiang, F.~Xiao, C.~Zhu, X.~Dai, D.~Wang,
  F.~Sun, F.~Iandola \emph{et~al.}, ``Efficientsam: Leveraged masked image
  pretraining for efficient segment anything,'' in \emph{Proceedings of the
  IEEE/CVF Conference on Computer Vision and Pattern Recognition}, 2024, pp.
  16\,111--16\,121.

\bibitem{kim2024otseg}
K.~Kim, Y.~Oh, and J.~C. Ye, ``Otseg: Multi-prompt sinkhorn attention for
  zero-shot semantic segmentation,'' in \emph{European Conference on Computer
  Vision}.\hskip 1em plus 0.5em minus 0.4em\relax Springer, 2024, pp. 200--217.

\bibitem{alexey2020image}
D.~Alexey, ``An image is worth 16x16 words: Transformers for image recognition
  at scale,'' \emph{arXiv preprint arXiv: 2010.11929}, 2020.

\bibitem{cong2024semi}
X.~Cong, J.~Gui, J.~Zhang, J.~Hou, and H.~Shen, ``A semi-supervised nighttime
  dehazing baseline with spatial-frequency aware and realistic brightness
  constraint,'' in \emph{Proceedings of the IEEE/CVF Conference on Computer
  Vision and Pattern Recognition}, 2024, pp. 2631--2640.

\bibitem{cordts2016cityscapes}
M.~Cordts, M.~Omran, S.~Ramos, T.~Rehfeld, M.~Enzweiler, R.~Benenson,
  U.~Franke, S.~Roth, and B.~Schiele, ``The cityscapes dataset for semantic
  urban scene understanding,'' in \emph{Proceedings of the IEEE conference on
  computer vision and pattern recognition}, 2016, pp. 3213--3223.

\bibitem{gupta2019lvis}
A.~Gupta, P.~Dollar, and R.~Girshick, ``Lvis: A dataset for large vocabulary
  instance segmentation,'' in \emph{Proceedings of the IEEE/CVF conference on
  computer vision and pattern recognition}, 2019, pp. 5356--5364.

\bibitem{qi2023small}
H.~Qi, H.~Zhou, J.~Dong, and X.~Dong, ``Small sample image segmentation by
  coupling convolutions and transformers,'' \emph{IEEE Transactions on Circuits
  and Systems for Video Technology}, vol.~34, no.~7, pp. 5282--5294, 2023.

\bibitem{wang2024polyp}
T.~Wang, X.~Qi, and G.~Yang, ``Polyp segmentation via semantic enhanced
  perceptual network,'' \emph{IEEE Transactions on Circuits and Systems for
  Video Technology}, vol.~34, no.~12, pp. 12\,594--12\,607, 2024.

\bibitem{chang2024DRNet}
Z.~Chang, X.~Gao, N.~Li, H.~Zhou, and Y.~Lu, ``Drnet: Disentanglement and
  recombination network for few-shot semantic segmentation,'' \emph{IEEE
  Transactions on Circuits and Systems for Video Technology}, vol.~34, no.~7,
  pp. 5560--5574, 2024.

\bibitem{wu2024toward}
J.~Wu, X.~Li, X.~Li, H.~Ding, Y.~Tong, and D.~Tao, ``Toward robust referring
  image segmentation,'' \emph{IEEE Transactions on Image Processing}, vol.~33,
  pp. 1782--1794, 2024.

\bibitem{nguyen2019multi}
K.~L. Nguyen, P.~Delachartre, and M.~Berthier, ``Multi-grid phase field skin
  tumor segmentation in 3d ultrasound images,'' \emph{IEEE Transactions on
  Image Processing}, vol.~28, no.~8, pp. 3678--3687, 2019.

\bibitem{gao2024efficient}
N.~Gao, X.~Jiang, X.~Zhang, and Y.~Deng, ``Efficient frequency-domain image
  deraining with contrastive regularization,'' in \emph{European Conference on
  Computer Vision}.\hskip 1em plus 0.5em minus 0.4em\relax Springer, 2024, pp.
  240--257.

\bibitem{hendrycks2016gaussian}
D.~Hendrycks and K.~Gimpel, ``Gaussian error linear units (gelus),''
  \emph{arXiv preprint arXiv:1606.08415}, 2016.

\bibitem{sandler2018mobilenetv2}
M.~Sandler, A.~Howard, M.~Zhu, A.~Zhmoginov, and L.-C. Chen, ``Mobilenetv2:
  Inverted residuals and linear bottlenecks,'' in \emph{Proceedings of the IEEE
  conference on computer vision and pattern recognition}, 2018, pp. 4510--4520.

\bibitem{kang2024metaseg}
B.~Kang, S.~Moon, Y.~Cho, H.~Yu, and S.-J. Kang, ``Metaseg: Metaformer-based
  global contexts-aware network for efficient semantic segmentation,'' in
  \emph{Proceedings of the IEEE/CVF winter conference on applications of
  computer vision}, 2024, pp. 434--443.

\bibitem{li2023scconv}
J.~Li, Y.~Wen, and L.~He, ``Scconv: Spatial and channel reconstruction
  convolution for feature redundancy,'' in \emph{Proceedings of the IEEE/CVF
  conference on computer vision and pattern recognition}, 2023, pp. 6153--6162.

\bibitem{hinton2006fast}
G.~E. Hinton, S.~Osindero, and Y.-W. Teh, ``A fast learning algorithm for deep
  belief nets,'' \emph{Neural computation}, vol.~18, no.~7, pp. 1527--1554,
  2006.

\bibitem{sudre2017generalised}
C.~H. Sudre, W.~Li, T.~Vercauteren, S.~Ourselin, and M.~Jorge~Cardoso,
  ``Generalised dice overlap as a deep learning loss function for highly
  unbalanced segmentations,'' in \emph{International Workshop on Deep Learning
  in Medical Image Analysis}.\hskip 1em plus 0.5em minus 0.4em\relax Springer,
  2017, pp. 240--248.

\bibitem{ke2023segment}
L.~Ke, M.~Ye, M.~Danelljan, Y.-W. Tai, C.-K. Tang, F.~Yu \emph{et~al.},
  ``Segment anything in high quality,'' \emph{Advances in Neural Information
  Processing Systems}, vol.~36, pp. 29\,914--29\,934, 2023.

\bibitem{shi2024zero}
Y.~Shi, D.~Liu, L.~Zhang, Y.~Tian, X.~Xia, and X.~Fu, ``Zero-ig: Zero-shot
  illumination-guided joint denoising and adaptive enhancement for low-light
  images,'' in \emph{Proceedings of the IEEE/CVF conference on computer vision
  and pattern recognition}, 2024, pp. 3015--3024.

\bibitem{liao2018hdpnet}
Y.~Liao, Z.~Su, X.~Liang, and B.~Qiu, ``Hdp-net: Haze density prediction
  network for nighttime dehazing,'' in \emph{Pacific Rim Conference on
  Multimedia}.\hskip 1em plus 0.5em minus 0.4em\relax Springer, 2018, pp.
  469--480.

\bibitem{wang2017learning}
L.~Wang, H.~Lu, Y.~Wang, M.~Feng, D.~Wang, B.~Yin, and X.~Ruan, ``Learning to
  detect salient objects with image-level supervision,'' in \emph{2017 IEEE
  conference on computer vision and pattern recognition (CVPR)}.\hskip 1em plus
  0.5em minus 0.4em\relax IEEE, 2017, pp. 3796--3805.

\bibitem{liew2021deep}
J.~H. Liew, S.~Cohen, B.~Price, L.~Mai, and J.~Feng, ``Deep interactive thin
  object selection,'' in \emph{Proceedings of the IEEE/CVF winter conference on
  applications of computer vision}, 2021, pp. 305--314.

\bibitem{zhou2023edgesam}
C.~Zhou, X.~Li, C.~C. Loy, and B.~Dai, ``Edgesam: Prompt-in-the-loop
  distillation for on-device deployment of sam,'' \emph{arXiv preprint
  arXiv:2312.06660}, 2023.

\bibitem{bradski2000opencv}
G.~Bradski, ``The opencv library.'' \emph{Dr. Dobb's Journal: Software Tools
  for the Professional Programmer}, vol.~25, no.~11, pp. 120--123, 2000.

\bibitem{qin2022highly}
X.~Qin, H.~Dai, X.~Hu, D.-P. Fan, L.~Shao, and L.~Van~Gool, ``Highly accurate
  dichotomous image segmentation,'' in \emph{European Conference on Computer
  Vision}.\hskip 1em plus 0.5em minus 0.4em\relax Springer, 2022, pp. 38--56.

\bibitem{li2020fss}
X.~Li, T.~Wei, Y.~P. Chen, Y.-W. Tai, and C.-K. Tang, ``Fss-1000: A 1000-class
  dataset for few-shot segmentation,'' in \emph{Proceedings of the IEEE/CVF
  conference on computer vision and pattern recognition}, 2020, pp. 2869--2878.

\bibitem{shi2015hierarchical}
J.~Shi, Q.~Yan, L.~Xu, and J.~Jia, ``Hierarchical image saliency detection on
  extended cssd,'' \emph{IEEE transactions on pattern analysis and machine
  intelligence}, vol.~38, no.~4, pp. 717--729, 2015.

\bibitem{cheng2014global}
M.-M. Cheng, N.~J. Mitra, X.~Huang, P.~H. Torr, and S.-M. Hu, ``Global contrast
  based salient region detection,'' \emph{IEEE transactions on pattern analysis
  and machine intelligence}, vol.~37, no.~3, pp. 569--582, 2014.

\bibitem{zeng2019towards}
Y.~Zeng, P.~Zhang, J.~Zhang, Z.~Lin, and H.~Lu, ``Towards high-resolution
  salient object detection,'' in \emph{Proceedings of the IEEE/CVF
  international conference on computer vision}, 2019, pp. 7234--7243.

\bibitem{zou2023generalized}
X.~Zou, Z.-Y. Dou, J.~Yang, Z.~Gan, L.~Li, C.~Li, X.~Dai, H.~Behl, J.~Wang,
  L.~Yuan \emph{et~al.}, ``Generalized decoding for pixel, image, and
  language,'' in \emph{Proceedings of the IEEE/CVF conference on computer
  vision and pattern recognition}, 2023, pp. 15\,116--15\,127.

\bibitem{paszke2019pytorch}
A.~Paszke, S.~Gross, F.~Massa, A.~Lerer, J.~Bradbury, G.~Chanan, T.~Killeen,
  Z.~Lin, N.~Gimelshein, L.~Antiga \emph{et~al.}, ``Pytorch: An imperative
  style, high-performance deep learning library,'' \emph{Advances in neural
  information processing systems}, vol.~32, 2019.

\bibitem{zhao2023unleashing}
W.~Zhao, Y.~Rao, Z.~Liu, B.~Liu, J.~Zhou, and J.~Lu, ``Unleashing text-to-image
  diffusion models for visual perception,'' in \emph{Proceedings of the
  IEEE/CVF International Conference on Computer Vision}, 2023, pp. 5729--5739.

\bibitem{cheng2021per}
B.~Cheng, A.~Schwing, and A.~Kirillov, ``Per-pixel classification is not all
  you need for semantic segmentation,'' \emph{Advances in neural information
  processing systems}, vol.~34, pp. 17\,864--17\,875, 2021.

\bibitem{liu2024grounding}
S.~Liu, Z.~Zeng, T.~Ren, F.~Li, H.~Zhang, J.~Yang, Q.~Jiang, C.~Li, J.~Yang,
  H.~Su \emph{et~al.}, ``Grounding dino: Marrying dino with grounded
  pre-training for open-set object detection,'' in \emph{European conference on
  computer vision}.\hskip 1em plus 0.5em minus 0.4em\relax Springer, 2024, pp.
  38--55.

\bibitem{zhao2023fast}
X.~Zhao, W.~Ding, Y.~An, Y.~Du, T.~Yu, M.~Li, M.~Tang, and J.~Wang, ``Fast
  segment anything,'' \emph{arXiv preprint arXiv:2306.12156}, 2023.

\bibitem{shu2025tinysam}
H.~Shu, W.~Li, Y.~Tang, Y.~Zhang, Y.~Chen, H.~Li, Y.~Wang, and X.~Chen,
  ``Tinysam: Pushing the envelope for efficient segment anything model,'' in
  \emph{Proceedings of the AAAI Conference on Artificial Intelligence},
  vol.~39, no.~19, 2025, pp. 20\,470--20\,478.

\bibitem{wang2023repvit}
A.~Wang, H.~Chen, Z.~Lin, J.~Han, and G.~Ding, ``Repvit-sam: Towards real-time
  segmenting anything,'' \emph{arXiv preprint arXiv:2312.05760}, 2023.

\bibitem{ravi2025sam2}
\BIBentryALTinterwordspacing
N.~Ravi, V.~Gabeur, Y.-T. Hu, R.~Hu, C.~Ryali, T.~Ma, H.~Khedr, R.~R{\"a}dle,
  C.~Rolland, L.~Gustafson, E.~Mintun, J.~Pan, K.~V. Alwala, N.~Carion, C.-Y.
  Wu, R.~Girshick, P.~Dollar, and C.~Feichtenhofer, ``{SAM} 2: Segment anything
  in images and videos,'' in \emph{The Thirteenth International Conference on
  Learning Representations}, 2025. [Online]. Available:
  \url{https://openreview.net/forum?id=Ha6RTeWMd0}
\BIBentrySTDinterwordspacing

\bibitem{hu2018squeeze}
J.~Hu, L.~Shen, and G.~Sun, ``Squeeze-and-excitation networks,'' in
  \emph{Proceedings of the IEEE conference on computer vision and pattern
  recognition}, 2018, pp. 7132--7141.

\bibitem{wang2020eca}
Q.~Wang, B.~Wu, P.~Zhu, P.~Li, W.~Zuo, and Q.~Hu, ``Eca-net: Efficient channel
  attention for deep convolutional neural networks,'' in \emph{Proceedings of
  the IEEE/CVF conference on computer vision and pattern recognition}, 2020,
  pp. 11\,534--11\,542.

\bibitem{woo2018cbam}
S.~Woo, J.~Park, J.-Y. Lee, and I.~S. Kweon, ``Cbam: Convolutional block
  attention module,'' in \emph{Proceedings of the European conference on
  computer vision (ECCV)}, 2018, pp. 3--19.

\end{thebibliography}

\begin{IEEEbiography}[{\includegraphics[width=1in,height=1.25in,clip,keepaspectratio]{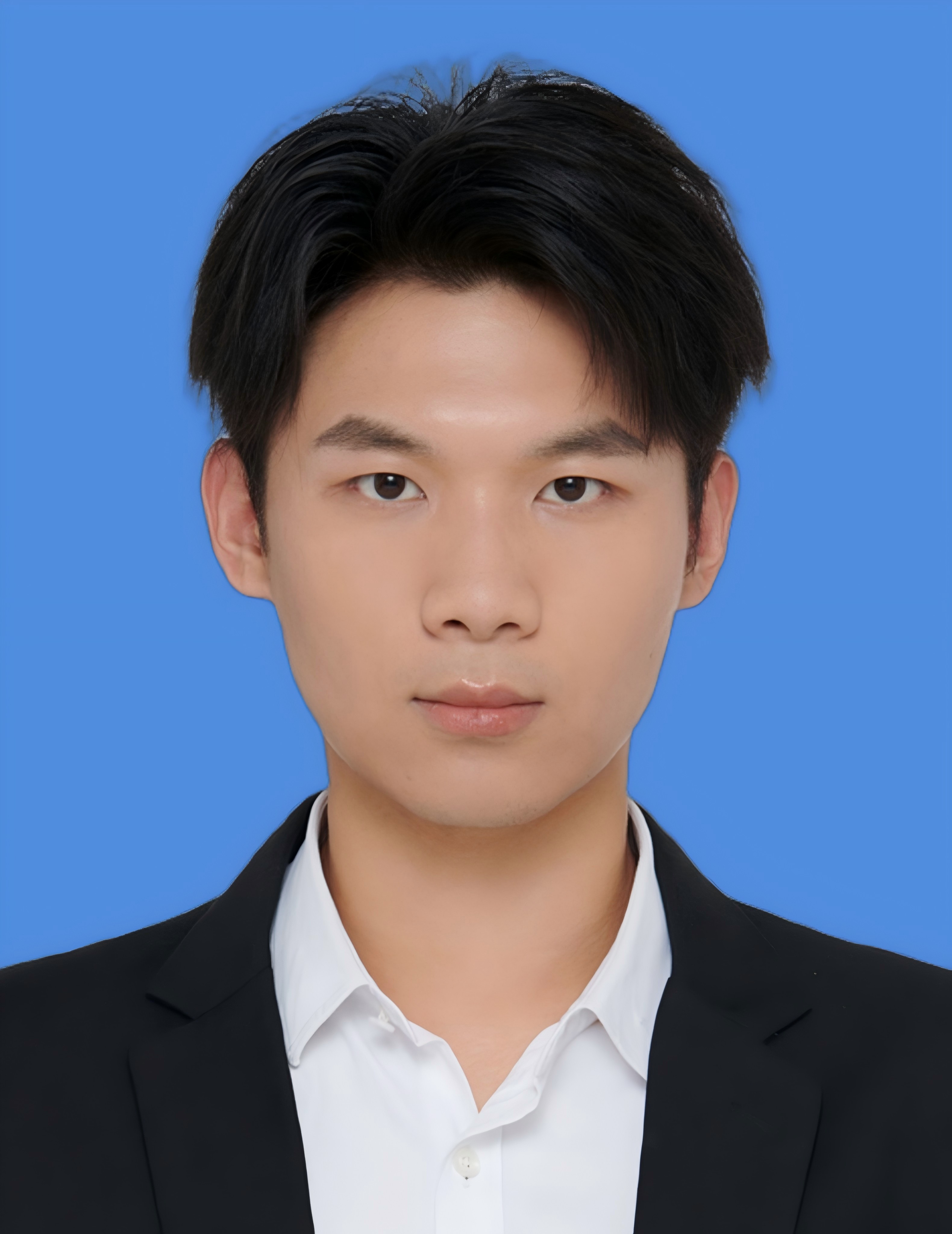}}]{Ruibo~Wang}
is currently pursuing the B.S. degree in Computer Science and Engineering with Delft University of Technology, Delft, Netherlands. Prior to this, he studied Computer Science and Technology at Hebei University of Technology. His research interests include computer vision, medical image analysis, and deep learning, with a specific focus on foundation models, image segmentation, and physics-guided image synthesis.
\end{IEEEbiography}

\begin{IEEEbiography}[{\includegraphics[width=1in,height=1.25in,clip,keepaspectratio]{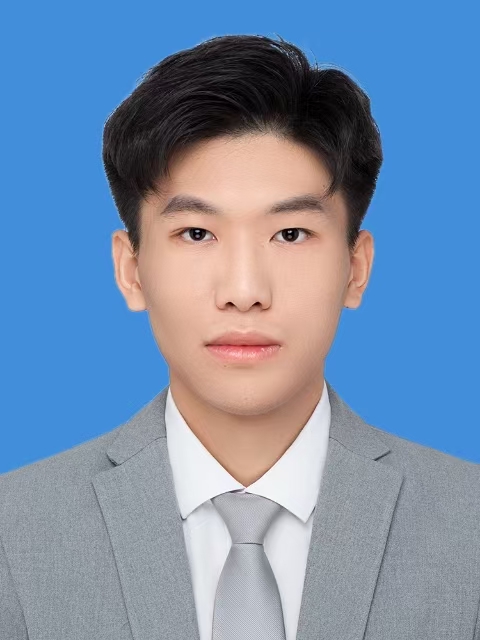}}]{Ziyi~Shen}
received the B.E. degree in medical information engineering from Jiangxi University of Chinese Medicine, Nanchang, China, in 2023. He is currently pursuing the master's degree in biomedical engineering with the School of Biomedical Engineering, Southern Medical University, Guangzhou, China. His research interests include medical image analysis and deep learning.
\end{IEEEbiography}

\begin{IEEEbiography}[{\includegraphics[width=1in,height=1.25in,clip,keepaspectratio]{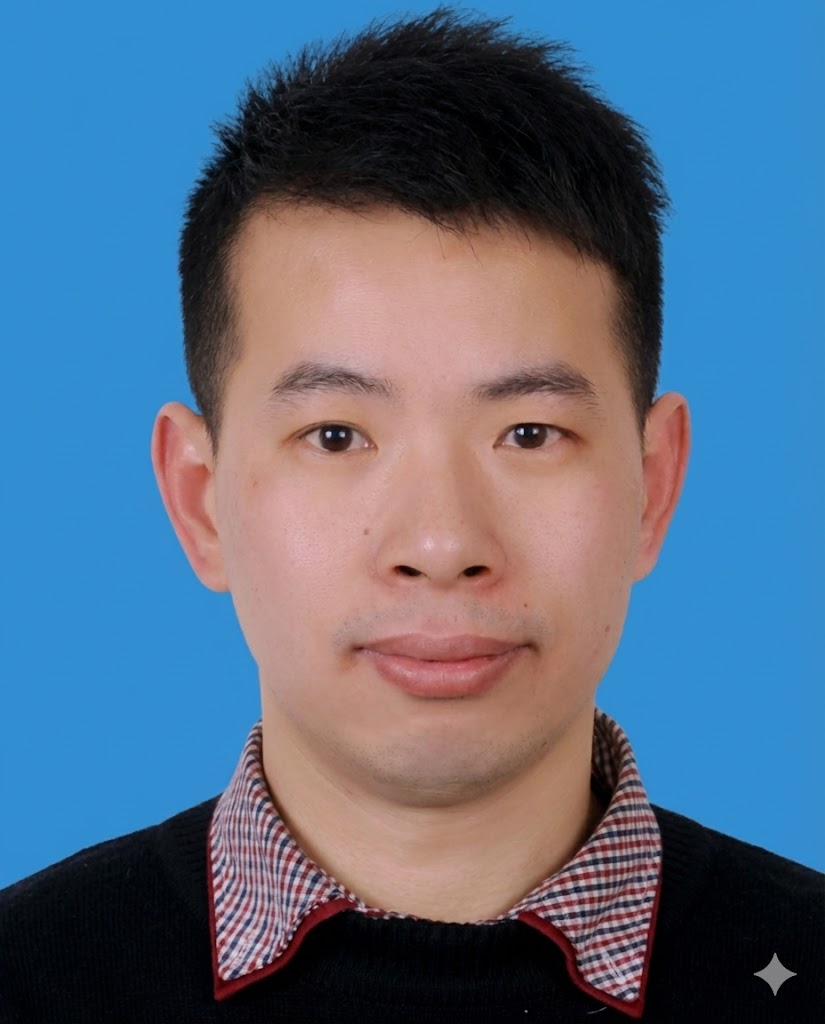}}]{Huaming~Wu}
(Senior Member, IEEE) received the BE and MS degrees in electrical engineering from the Harbin Institute of Technology, China, in 2009 and 2011, respectively, and the PhD degree with the highest honor in computer science from Freie Universit\"at Berlin, Germany, in 2015. He is currently a professor with the Center for Applied Mathematics, Tianjin University, China. His research interests include mobile cloud computing, edge computing, internet of things, deep learning, complex networks, and DNA storage. He currently serves as an Associate Editor for several IEEE journals, including IEEE Transactions on Dependable and Secure Computing, IEEE Transactions on Intelligent Transportation Systems, IEEE Transactions on Circuits and Systems for Video Technology, IEEE Transactions on Consumer Electronics, and IEEE Transactions on Technology and Society.
\end{IEEEbiography}

\begin{IEEEbiography}[{\includegraphics[width=1in,height=1.25in,clip,keepaspectratio]{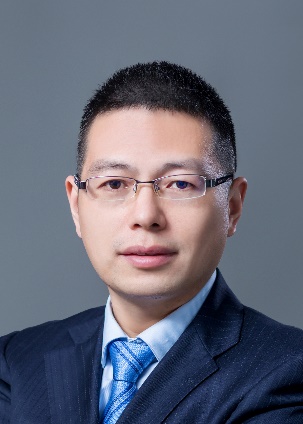}}]{Dong~Liang}
(Senior Member, IEEE) received the bachelor's degree in electronics engineering and the M.S. degree in signal and information processing from the Hefei University of Technology, Hefei, China, in 1998 and 2002, respectively, and the Ph.D. degree in pattern recognition and intelligent system from Shanghai Jiao Tong University, Shanghai, China, in 2006. He is currently a Professor with the Shenzhen Institutes of Advanced Technology, Chinese Academy of Sciences, Shenzhen, China. From 2007 and 2011, he conducted postdoctoral research with the University of Hong Kong, Hong Kong, and with the University of Wisconsin-Milwaukee, Milwaukee, WI, USA. He is currently working on signal processing, machine learning, and biomedical imaging. He currently acts as an editorial member of the international journal IEEE Transactions on Medical Imaging and Magnetic Resonance in Medicine.
\end{IEEEbiography}

\begin{IEEEbiography}[{\includegraphics[width=1in,height=1.25in,clip,keepaspectratio]{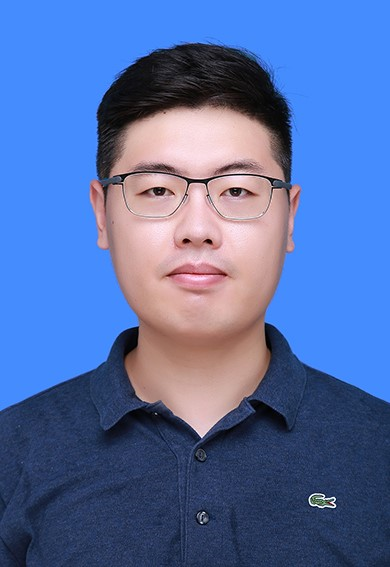}}]{Kun~Shang}
received the B.S. degree from the Faculty of mathematics and statistics, Hubei University, in 2011. In 2018, He received the M.S. and Ph.D. degrees from Center for Applied Mathematics, Tianjin University. From 2018 to 2021, he was an Assistant professor of Mathematics with the School of Mathematics, Hunan University. He is currently an Associate Professor with the Shenzhen Institutes of Advanced Technology, Chinese Academy of Sciences, Shenzhen, China. Recently, his research interests include Brain-inspired computing, Brain Computer Interface, optimization etc.
\end{IEEEbiography}

\vfill

\end{document}